\documentclass[letterpaper]{article} 
\usepackage{aaai2026}  
\usepackage{times}  
\usepackage{helvet}  
\usepackage{courier}  
\usepackage[hyphens]{url}  
\usepackage{graphicx} 
\usepackage{natbib}  
\usepackage{caption} 
\usepackage{algorithm}
\usepackage{algorithmic}
\usepackage{pifont}
\usepackage{newfloat}
\usepackage{listings}
\DeclareCaptionStyle{ruled}{labelfont=normalfont,labelsep=colon,strut=off} 
\floatstyle{ruled}
\newfloat{listing}{tb}{lst}{}
\floatname{listing}{Listing}
\usepackage{dsfont}
\usepackage{xcolor}
\usepackage[table]{xcolor}
\definecolor{mygray}{gray}{0.9}
\definecolor{lavender}{RGB}{230,230,250}
\definecolor{light}{HTML}{cae9ff}

\title{Clear Nights Ahead: Towards Multi-Weather Nighttime Image Restoration}
\author{
    Yuetong Liu\textsuperscript{\rm 1,3}, Yunqiu Xu\textsuperscript{\rm 4,}\thanks{Corresponding authors.}, Yang Wei\textsuperscript{\rm 2,3}, Xiuli Bi\textsuperscript{\rm 2,3}, Bin Xiao\textsuperscript{\rm 2,3,5,${\ast}$}\\
}
\affiliations{
    \textsuperscript{\rm 1}School of Computer Science and Technology, Chongqing University of Posts and Telecommunications\\
    \textsuperscript{\rm 2}School of Artificial Intelligence, Chongqing University of Posts and Telecommunications\\
    \textsuperscript{\rm 3}Chongqing Key Laboratory of Image Cognition, Chongqing University of Posts and Telecommunications\\
    \textsuperscript{\rm 4}ReLER Lab, CCAI, Zhejiang University \qquad
    \textsuperscript{\rm 5}Jinan Inspur Data Technology Co., Ltd.\\
    d230201022@stu.cqupt.edu.cn, imyunqiuxu@gmail.com,
    \{weiyang, bixl, xiaobin\}@cqupt.edu.cn
}
\usepackage{bibentry}

\begin{document}

\maketitle

\begin{abstract}
Restoring nighttime images affected by multiple adverse weather conditions is a practical yet under-explored research problem, as multiple weather degradations usually coexist in the real world alongside various lighting effects at night. 
This paper first explores the challenging multi-weather nighttime image restoration task, where various types of weather degradations are intertwined with flare effects.
To support the research, we contribute the AllWeatherNight dataset, featuring large-scale nighttime images with diverse compositional degradations.
By employing illumination-aware degradation generation, our dataset significantly enhances the realism of synthetic degradations in nighttime scenes, providing a more reliable benchmark for model training and evaluation.
Additionally, we propose ClearNight, a unified nighttime image restoration framework, which effectively removes complex degradations in one go.
Specifically, ClearNight extracts Retinex-based dual priors and explicitly guides the network to focus on uneven illumination regions and intrinsic texture contents respectively, thereby enhancing restoration effectiveness in nighttime scenarios. 
Moreover, to more effectively model the common and unique characteristics of multiple weather degradations, ClearNight performs weather-aware dynamic specificity and commonality collaboration that adaptively allocates optimal sub-networks associated with specific weather types.
Comprehensive experiments on both synthetic and real-world images demonstrate the necessity of the AllWeatherNight dataset and the superior performance of ClearNight.
\end{abstract}

\begin{links}
    \link{Project Page:}{https://henlyta.github.io/ClearNight/}
\end{links}

\section{Introduction}
\label{sec:intro}

\begin{figure}[ht!]
\centering
    \includegraphics[width=\linewidth]{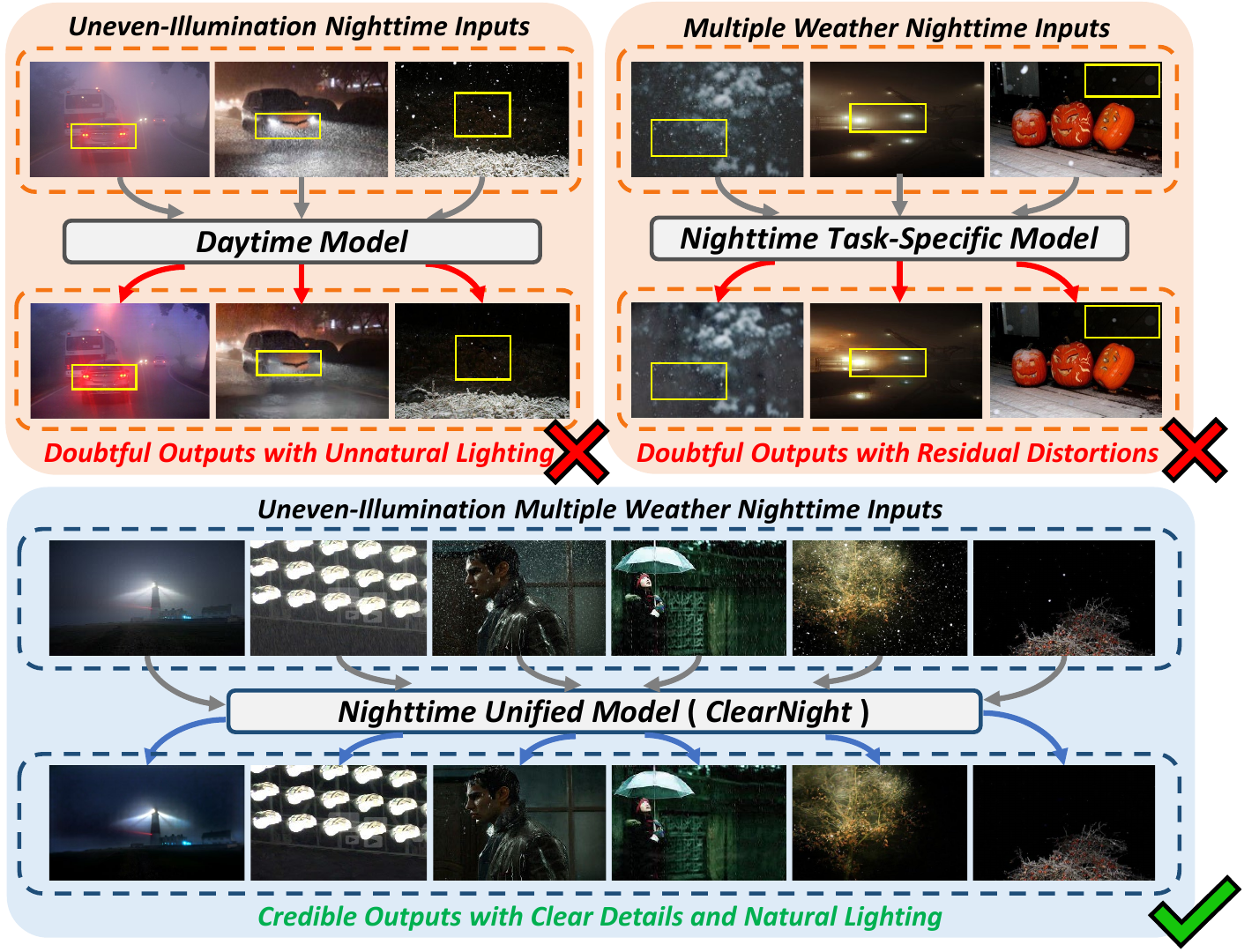}\\ 
\caption{ClearNight is the first multi-weather nighttime image restoration framework, which effectively removes complex and coupled weather and flare degradations in one go.}
\label{intro1}
\end{figure}

\begin{table}[t]
        \centering
        \small
        \setlength{\tabcolsep}{3.85pt}
        \begin{tabular}{l|ccccc|cc|c}
            \hline
            \rowcolor{mygray}\textbf{Datasets} & \textbf{H} & \textbf{RS} & \textbf{RD} & \textbf{S} & \textbf{F} & \textbf{Syn}& \textbf{Real} & \textbf{Night}\\     
            \hline
            Outdoor-Rain & \ding{52} & \ding{52} & \ding{56} & \ding{56} & \ding{56} &\ding{52} & \ding{56}& \ding{56} \\
            RainDS & \ding{56} & \ding{52} & \ding{52} & \ding{56}  & \ding{56} &\ding{52} & \ding{52}& \ding{56}\\ 
            BID-II & \ding{52} & \ding{52} & \ding{52} & \ding{52}  & \ding{56} &\ding{52} & \ding{56}& \ding{56}\\
            WeatherStream & \ding{52} & \ding{52} & \ding{52} & \ding{52} & \ding{56} &\ding{56} & \ding{52}& \ding{56}\\
            CDD-11& \ding{52} & \ding{52} & \ding{56} & \ding{52} &\ding{56}&\ding{52} & \ding{56}&\ding{56}\\
            \hline
            Raindrop Clarity & \ding{56} & \ding{56} & \ding{52} & \ding{56} & \ding{56} & \ding{56} & \ding{52} & \ding{52} \\
            Night/YellowHaze & \ding{52} & \ding{56} & \ding{56}  & \ding{56} & \ding{56} &\ding{52} & \ding{56} & \ding{52}\\
            NHC/NHM/NHR & \ding{52} & \ding{56} & \ding{56} & \ding{56} & \ding{56} &\ding{52} & \ding{52}& \ding{52}\\
            GTA5 & \ding{52} & \ding{56} & \ding{56} & \ding{56} & \ding{56} &\ding{52} & \ding{56}& \ding{52}\\
            UNREAL-NH & \ding{52} & \ding{56} & \ding{56} & \ding{56} & \ding{52} &\ding{52} & \ding{52}& \ding{52}\\ 
            GTAV-NightRain & \ding{56} & \ding{52} & \ding{56} & \ding{56} & \ding{56} &\ding{52} & \ding{56} & \ding{52}\\
            RVSD & \ding{56} & \ding{56} & \ding{56} & \ding{52} & \ding{56}  &\ding{52} & \ding{56} & \ding{52}\\
            AllWeatherNight & \ding{52} & \ding{52} & \ding{52} & \ding{52} & \ding{52} &\ding{52} & \ding{52}& \ding{52}\\
            \hline
        \end{tabular}
        \caption{Comparison to related adverse weather datasets. \textbf{H}, \textbf{RS}, \textbf{RD} and \textbf{S} indicate haze, rain streak, raindrop and snow, respectively. \textbf{F} denotes the presence of synthesized flare images. \textbf{Syn} denotes synthesized images. \textbf{Night} denotes the presence of nighttime data.}
        \label{datasets}
\end{table}

Image restoration under adverse weather conditions is a vital preprocessing step in many computer vision applications, such as autonomous driving~\cite{BDD100K,quan2021holistic,driving2,xu2024mc,xu2022h2fa} and video surveillance~\cite{liu2023generating,li2024,li2025,AAAI2026,wu2025bvinet}.
Although significant efforts have been made to tackle the issues posed by adverse weather images, prior works~\cite{HECCV2024,LCVPR2024,GOUBICML2024} often neglect the complexities of lighting conditions (\textit{e.g.}, flare effects) and their intricate interplay with weather degradations, limiting the effectiveness in real-world applications.

In particular, nighttime scenes exacerbate such problem, where adverse weather degradations and uneven illumination are tightly coupled, severely obscuring background contents.
Although a few works~\cite{NACM2020,NACM20232,NICCV2023} have explored nighttime adverse weather image restoration, they seldom take into account these unique characteristics.
More critically, existing methods are limited to handling a single type of degradations, which leads to unsatisfactory performance in complex real-world scenarios where multiple adverse weather conditions frequently co-occur (see Fig.~\ref{intro1}).
To better meet the demands of real-world applications, this paper first explores a highly practical yet remains largely under-explored task: multi-weather nighttime image restoration. 
In this context, two critical challenges must be addressed: \textbf{\ding{182} the scarcity of realistic multi-weather training samples} and \textbf{\ding{183} the insufficient ability of existing models to effectively address entangled degradations in nighttime scenes}.

\begin{figure}[t]
        \centering
        \includegraphics[width=1\linewidth]{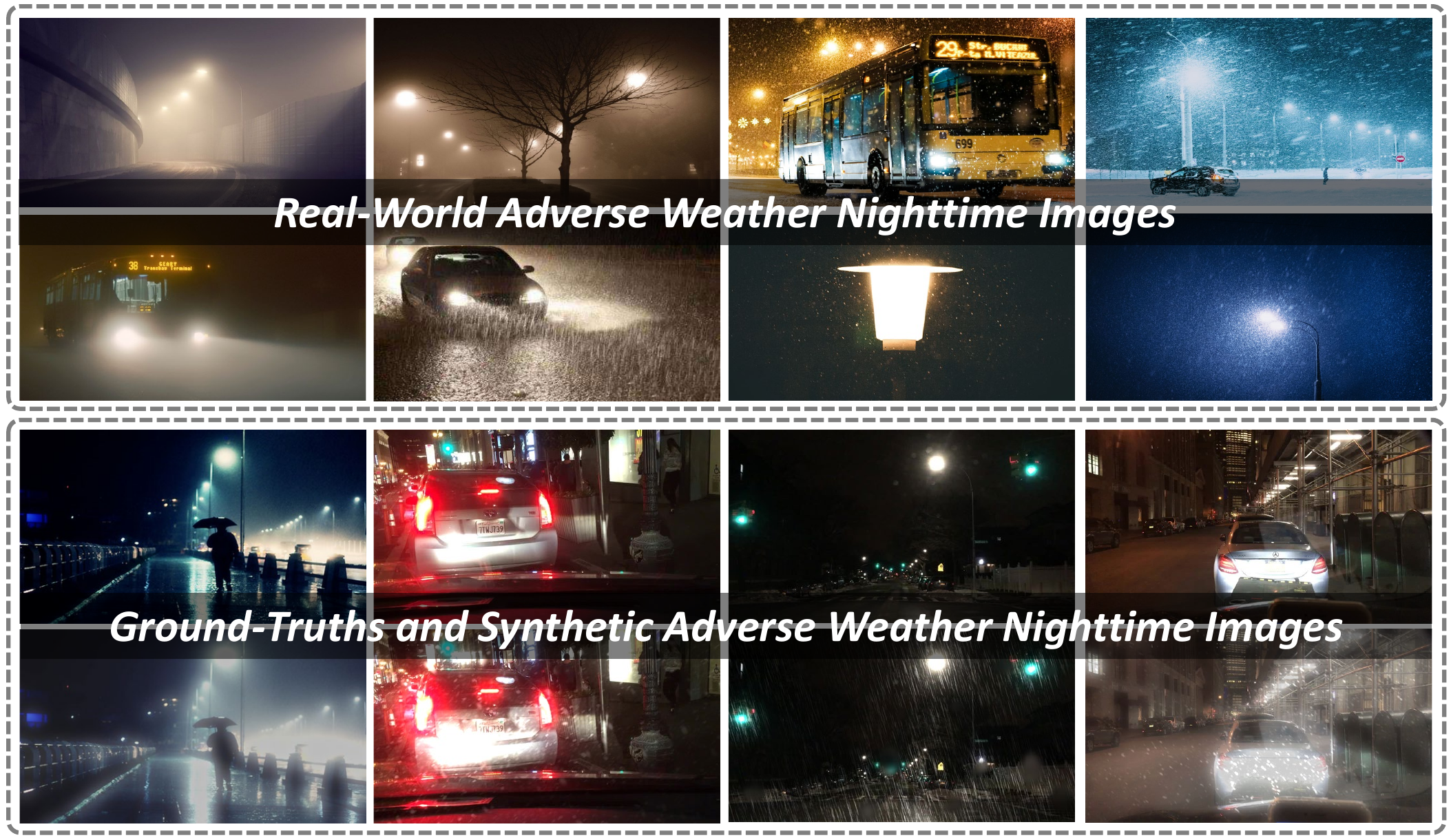}\\
        \captionof{figure}{Real-world and synthetic samples in our AllWeatherNight dataset. The synthetic images effectively simulate real-world nighttime scenes with various degradations.}
        \label{illum_region}
\end{figure}

To the best of our knowledge, there is no existing dataset that is applicable to multi-weather nighttime image restoration.
As summarized in Tab.~\ref{datasets}, all previous datasets~\cite{OneRestoreECCV2024,NRaindropECCV2024,NACM20232,NICCV2023,SICCV2023} overlook nighttime scenarios with multiple weather degradations under non-uniform illumination.
To facilitate the research, we construct AllWeatherNight, a dataset for restoring nighttime images affected by multiple adverse weather effects and flares (see Fig.~\ref{illum_region}).
Specifically, we collect diverse nighttime images from multiple sources and design an illumination-aware degradation generation method to faithfully emulate degradations in real-world nighttime photos, yielding images characterized by the realistic intertwined effects of uneven illumination and multiple adverse weather degradations.
Consequently, training with our synthesized images enables models to generalize better in real-world nighttime scenarios. 

To effectively handle the intertwined degradations caused by artificial lighting and multiple adverse weather effects at night, we present ClearNight, the first unified framework tailored for multi-weather nighttime image restoration.
We first leverage Retinex-based dual prior guidance to explicit disentangle the lighting and texture information within degraded images.
Specifically, the decoupled illumination prior guides the model to focus on the uneven lit regions for effective restoration in nighttime scenes, while the reflectance prior enhances texture representations to mitigate weather-induced degradations and recover clear background details.

In order to better represent complex degradations consisting of multiple adverse weather conditions, ClearNight employs a specificity-commonality dual-branched architecture, where the specificity branch is dynamically constructed and synergizes with the commonality branch.
The dynamic specificity branch contains diverse sub-networks, formed by selecting different combinations of candidate units~\cite{losslb,specific,jia2024mos2,llm2024}.
To further enhance the capabilities of handling multi-weather degradations, we associate these candidate units with specific weather types using an auxiliary weather instructor, which identifies weather degradations and implicitly guides their dynamic allocation.
Thus, the dynamic allocator becomes weather-aware and is encouraged to consistently select appropriate units for the same types of weather conditions.

Extensive experiments and analysis demonstrate the superiority of our ClearNight on both synthetic and real-world images.
The main contributions are summarized as follows:

\begin{itemize}
\item This work pioneers multi-weather image restoration in nighttime conditions. We contribute a new dataset featuring 10K realistic illumination-aware synthetic images with multi-degradation alongside real-world samples.

\item We propose ClearNight, the first unified framework for multi-weather nighttime image restoration, which integrates Retinex-based dual prior guidance and weather-aware dynamic specificity-commonality collaboration, tailored to address challenging uneven lighting and diverse weather effects entangled in nighttime scenes.

\item Our ClearNight effectively removes various degradations and outperforms state-of-the-art approaches on both synthetic and real-world adverse weather nighttime images.

\end{itemize}

\section{Related Work}
\label{sec:Related Work}
\noindent\textbf{Adverse Weather Nighttime Image Restoration.}
Early works~\cite{NICIP2014,NCVPR2017} typically relied on physical priors and statistical assumptions, limiting their effectiveness and robustness in handling real-world scenarios.
To address the limitations, several data-driven methods have been developed, achieving impressive results in restoring nighttime images degraded by haze~\cite{NICIP2012,NACM20232,NCVPR2024,NTIP2022,retinex7,CVPRW2022,NAAAI2025}, 
rain~\cite{NICCV2023,NAAAI2024} 
and snow~\cite{SICCV2023}. 
However, existing works focus on single-type degradations and overlook the fact that real-world nighttime scenes often involve multiple simultaneous degradations caused by adverse weather and uneven lighting.

\noindent\textbf{Multi-Weather Image Restoration.}
Multi-weather image restoration aims to restore various weather-degraded scenes using a single model~\cite{UMWECCV2022,FewECCV2024,LCVPR2024,MPPCVPR2024,LAAAI2024,RAAAI2024,DAAAI2025,UICLR2025,TSD2025}.
Recently, numerous Transformer-based approaches are developed, which investigate weather queries~\cite{TransCVPR2022}, adverse weather pixel categorization~\cite{HistogramECCV2024}, texture-guided appearance flow~\cite{CPCVPR2023}, codebook prior fusion~\cite{AWICCV2023}, and \textit{etc}.
Moreover, several recent works~\cite{WeatherDiff2023,TTTECCV2024,SCVPR2024} leverage remarkable generative capabilities of diffusion models for restoration performance boosting. 
To the best of our knowledge, all the previous methods focus on daytime scenes and ignore the entanglement of multiple weather conditions with non-uniform lighting in nighttime scenes.

\noindent\textbf{Adverse Weather Datasets.}
Most previous datasets, like Outdoor-Rain~\cite{HCVPR2019}, RainDS~\cite{GOCVPR2021}, BID-II~\cite{BECCV2022}, CDD-11~\cite{OneRestoreECCV2024} and WeatherStream~\cite{WSCVPR2023}, only focus on adverse weather image restoration in daytime scenarios.
A few recent studies explore this task in nighttime scenes and build datasets for haze (\textit{e.g.} Night/YellowHaze~\cite{NPCM2018}, NHC/NHM/NHR~\cite{NACM2020}, UNREAL-NH~\cite{NACM20232}), rain (\textit{e.g.} GTA5~\cite{NECCV2020}, GTAV-NightRain~\cite{GTAV,NICCV2023}, Raindrop Clarity~\cite{NRaindropECCV2024}) and snow (\textit{e.g.} RVSD~\cite{SICCV2023}).
However, each existing nighttime dataset only considers a single type of weather condition, ignoring the challenge of mixed weather conditions commonly found in the real world.

\section{AllWeatherNight Dataset}\label{sec:Datasets}

Real-world nighttime scenes are often degraded by a complex interplay of co-occurring weather and flares, challenging models for faithful restoration and visibility enhancement. To address this gap, we introduce AllWeatherNight, a new large-scale dataset featuring nighttime images with coupled degradations from mixed weather and uneven lighting.

\begin{figure}[t]
\centering   
\includegraphics[width=0.95\linewidth]{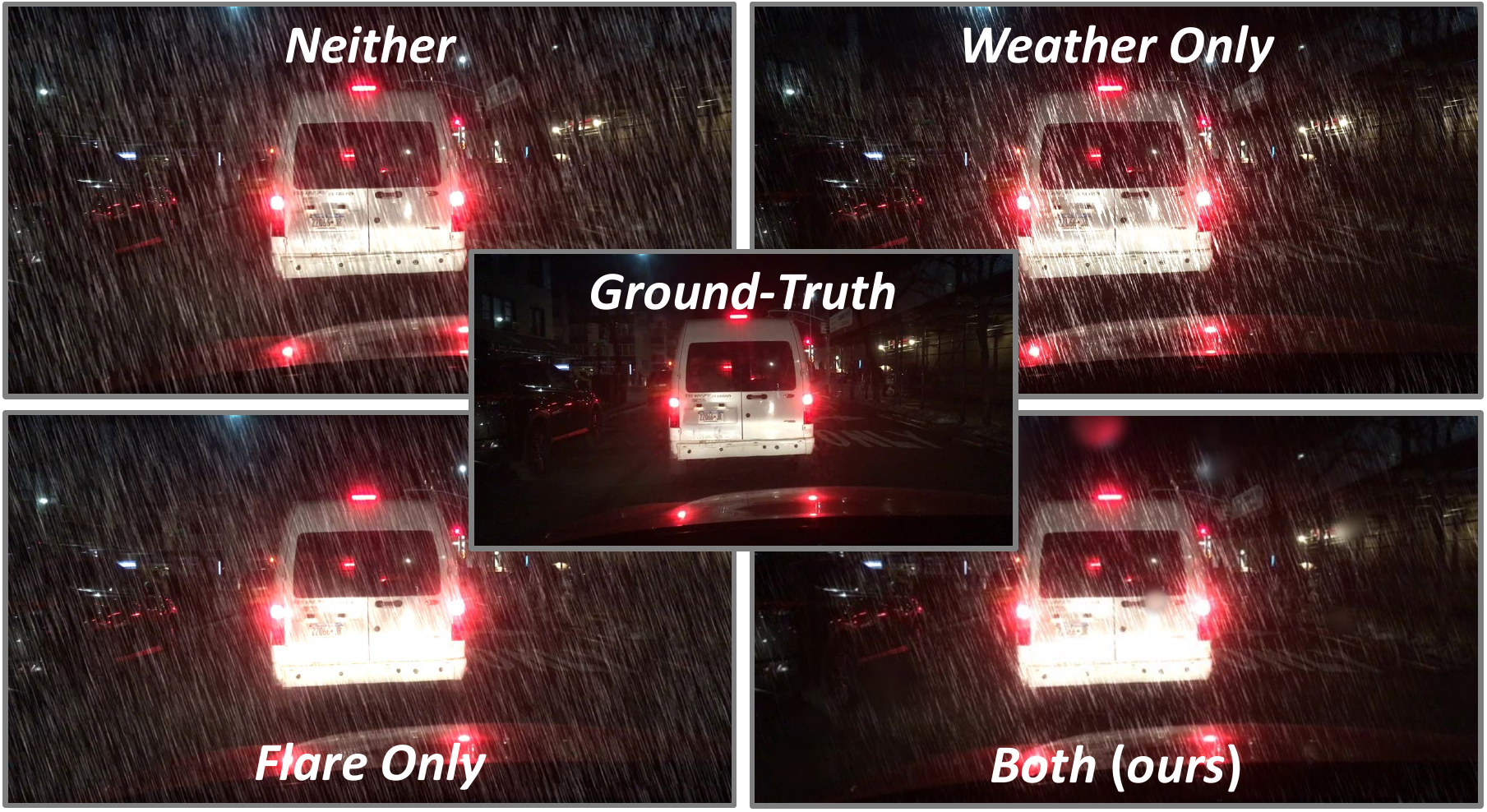}
\caption{
Visualization of four synthesized variants of complex rain scene, where \textbf{Weather Only} and \textbf{Flare Only} denote synthesis with illumination-aware weather degradation and flare, respectively. Ours involves both degradations. }
\label{visual_dataset}
\end{figure}

\noindent\textbf{Data Collection and Filtering.}
We repurpose images from BDD100K~\cite{BDD100K} and ExDark~\cite{Exdark} as ground truths.
These candidate images undergo a two-step process: 
selecting high-quality nighttime samples using average brightness, average gradient and grayscale variance; and manually selecting the 1,000 diverse images from each dataset.
We also collect 1,000 real-world nighttime images with various weather conditions from the Internet and existing datasets~\cite{RESIDE,SPA,snow100k,GAN,DDN,NACM2023,JORDER}.

\begin{figure}[t!]
\centering
\includegraphics[width=1\linewidth]{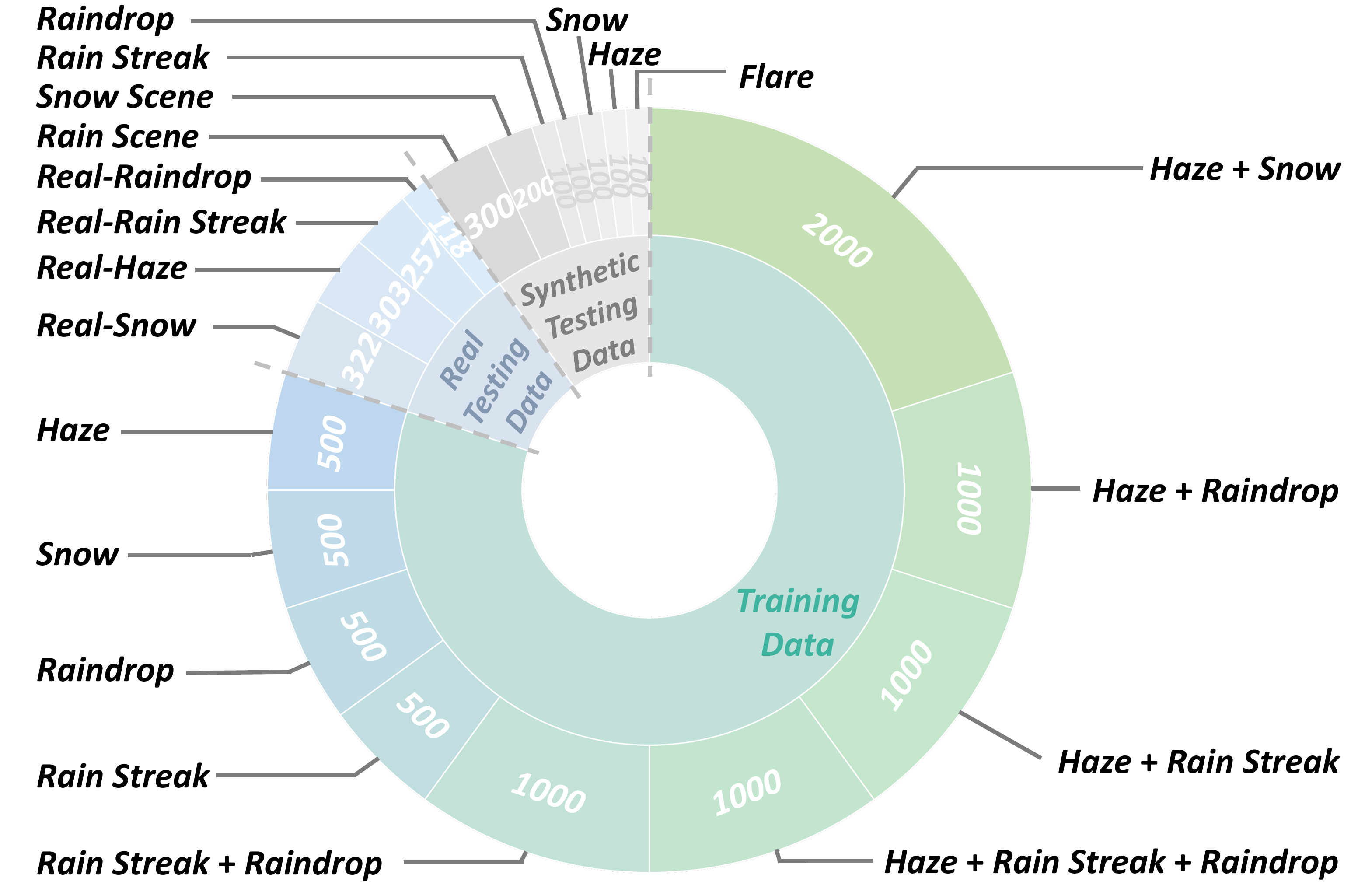}
\caption{Distribution of the our AllWeatherNight dataset.}
\label{dataset_num}
\end{figure}

\begin{figure*}[t]
\centering
\includegraphics[width=\linewidth]{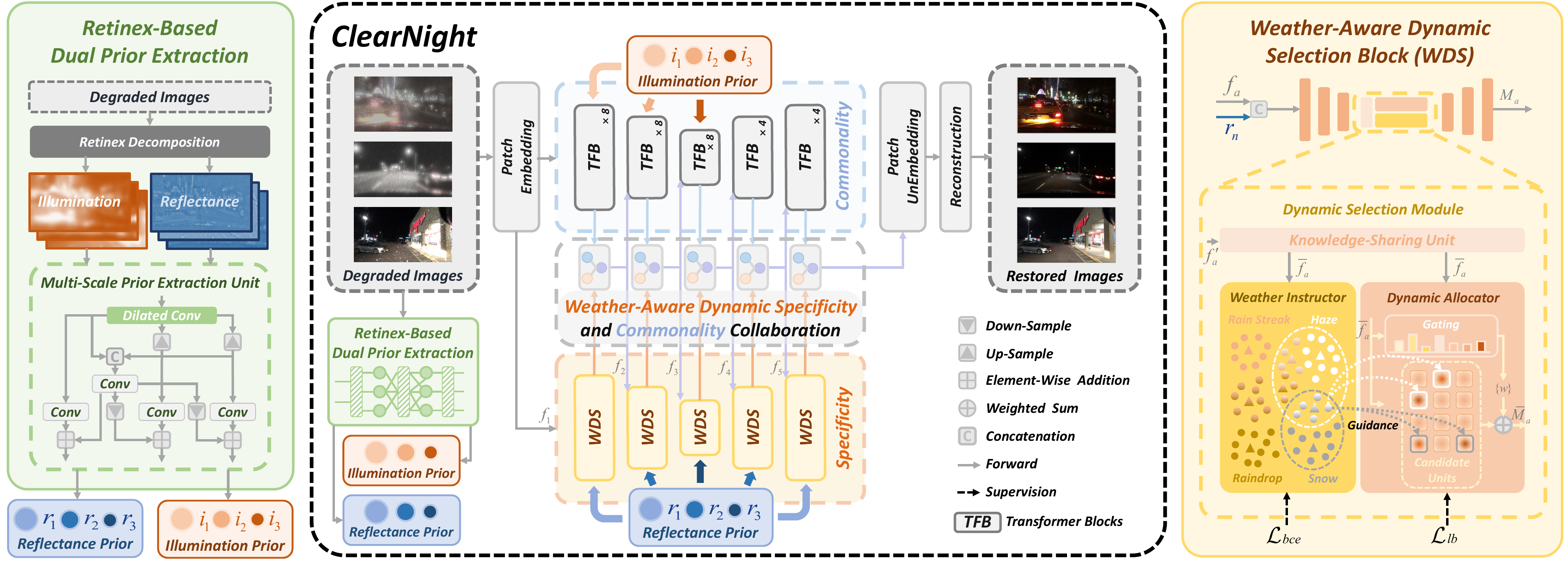}
\caption{Overview of our ClearNight framework. ClearNight primarily comprises Retinex-based dual prior guidance as well as weather-aware dynamic specificity and commonality branches. 
The Retinex-based dual priors explicitly guide the network to focus on illumination regions and intrinsic textures.
The weather-aware dynamic specificity branch adaptively accommodates various weather effects and collaborates with the commonality branch to effectively handle complex multi-weather scenes.}
\label{frame}
\end{figure*}

\noindent\textbf{Illumination-Aware Degradation Generation.}
We observe that uneven lighting conditions in real-world nighttime scenes often interact with adverse weather degradations, yet this phenomenon is largely overlooked by prior works~\cite{NACM2020,NICCV2023}.
To synthesize more realistic nighttime images with weather effects, we introduce an illumination-aware degradation generation approach that first modulates the lighting effects and then superimposes an illumination-aware combination of weather degradations.

Given a clean nighttime image $\mathit{X}$, we generate flare-degraded image $\mathit{X}^\mathrm{flare}$ based on light regions:
\begin{equation}
\mathit{X}^\mathrm{flare} = \alpha \cdot \mathit{X} + \beta \cdot (\mathit{L} \ast \mathit{K}^\mathrm{APSF}),
\end{equation}
where $\alpha$ controls clean image preservation. $\beta$ regulates flare blending and is adaptively set by the light source pixel ratio.
$\mathit{L}$ is a light source map devised via thresholding and alpha matting refinement~\cite{alpha}.
$\mathit{K}^\mathrm{APSF}$ is a 2D kernel from the atmospheric point spread function~\cite{NACM2023}, and $\ast$ is the convolution operator.

Subsequently, building upon $\mathit{X}^\mathrm{flare}$, the illumination-aware weather-degraded image $\mathit{X}^\mathrm{d}$ is synthesized by:
\begin{equation}
\mathit{X}^\mathrm{d} = \mathit{X}^\mathrm{flare} + \sum_{e\in \mathcal{E}}\omega_e \cdot \mathcal{F}^\mathrm{G}_e (\mathit{X}^\mathrm{flare}),
\end{equation}
where $\mathcal{E}$ is the set of selected weather effects for generating a specific degraded image, which is a subset of the universal set of all possible effects, \textit{i.e.}, $\mathcal{E} \subseteq \{\mathrm{H}, \mathrm{RS}, \mathrm{RD}, \mathrm{S}\}$ (Haze, Rain Streak, Raindrop, and Snow).
We devise a weight map $\omega_e$ to better simulate weather degradations under uneven lighting, where $\omega_{e\neq\mathrm{RD}}$ is set to the Retinex decomposed illumination map, while $\omega_{e=\mathrm{RD}}$ is empirically set to 1, as raindrops are primarily affected by local background rather than distant lighting.
$\mathcal{F}^\mathrm{G}_e(\cdot)$ represents corresponding weather generation functions.
We simulate adverse weather using established models~\cite{HCVPR2019,Raindrop,snow100k,csd}.
As shown in Fig.~\ref{visual_dataset}, unlike simplistic, spatially uniform synthesis, illumination-aware method generates more natural degradations. 

\noindent\textbf{Dataset Statistics.}
Overall, we synthesize 8,000 nighttime training images, covering multiple weather with varying scales, directions, patterns and intensities.
As summarized in Fig.~\ref{dataset_num}, the test dataset comprises synthetic and real-world subsets, each containing 1,000 images for model evaluation.

\section{ClearNight}
\label{sec:Method}

ClearNight is a unified nighttime image restoration framework designed to simultaneously remove multiple weather degradations.
As depicted in Fig.~\ref{frame}, ClearNight integrates Retinex-based dual prior guidance and weather-aware dynamic specificity-commonality collaboration. 
The former explicitly decouples uneven lighting and textures to guide the network in recovering clear images with natural lighting and rich background details.
The latter effectively captures the unique and shared characteristics of diverse weather conditions, enabling powerful multi-weather image restoration.

\subsection{Retinex-Based Dual Prior Guidance}
Decoupling illumination and texture information is critical for nighttime image restoration, as it allows the model to separately handle non-uniform lighting and adverse weather degradations.
Retinex theory~\cite{retinex1} provides a classic physical model for this decomposition, formulating the input degraded image $\mathit{X}^\mathrm{d}$ as:
\begin{equation}
\mathit{X}^\mathrm{d} = \mathit{R}^\mathrm{d} \cdot \mathit{I}^\mathrm{d},
\end{equation}
where $\mathit{R}^\mathrm{d}$ and $\mathit{I}^\mathrm{d}$ are the reflectance and illumination components, respectively. The decomposed components $\mathit{p}^\mathrm{rtx}\in\{\mathit{I}^\mathrm{d}, \mathit{R}^\mathrm{d}\}$ are then fed into shared-weight multi-scale prior extraction units (MPE):
\begin{equation}
   \mathit{p}_1, \mathit{p}_2, \mathit{p}_3 = \mathcal{F}^\mathrm{MPE} (\mathit{p}^\mathrm{rtx}) \quad \textrm{with} \quad \mathit{p}_n \in \{\mathit{i}_n, \mathit{r}_n\},
\end{equation}
where $\mathit{p}_n$ denotes the illumination or reflectance prior at the $n$-th scale, with $n\in\{1, 2, 3\}$. 
Within each MPE unit, a dilated convolution first projects the $\mathit{p}^\mathit{rtx}$ into three scales. 
Then, resulting multi-scale features are interactively fused to produce the final illumination/reflectance priors $\mathit{p}_n$.

More specifically, the illumination priors $\mathit{i}_n$ are successively injected into the Transformer blocks (TFBs) of the first three stages, which guide the network to focus on the uneven lighting regions in nighttime images, thereby facilitating the handling of lighting-influenced weather degradations.
As the reflectance priors $\mathit{r}_n$ contain rich intrinsic textures that not only capture background details but also reveal degradation types, we incorporate them into each weather-aware dynamic selection block (WDS) to enhance weather type discrimination and improve multi-weather restoration.

\subsection{Weather-Aware Dynamic Specificity-Commonality Collaboration}
\noindent\textbf{Dynamic Specificity and Commonality Synergy.}
As different weather degradations (\textit{e.g.}, snow and rain streaks) exhibit both shared and unique patterns, we introduce a synergistic design of dynamic specificity and commonality to effectively model complex multi-weather degradations. 
The commonality branch consists of sequential Transformer blocks (TFBs), while the specificity branch incorporates multiple weather-aware dynamic selection blocks (WDS) as residuals for each stage of TFBs.

As shown in the right of Fig.~\ref{frame}, each WDS consists of an encoder-decoder structure and a dynamic selection module, which jointly process the merged features from two branches and the reflectance prior.
The encoder-decoder structure, inspired by \cite{specific}, learns compact representations,
while the dynamic selection module constructs input-tailored sub-networks by sparsely selecting candidate units ($\mathcal{F}^\mathrm{U}_k(\cdot)$), where $k$ is the unit index.

Besides the candidate units, the dynamic selection module includes a gating ($\mathcal{F}^\mathrm{W}(\cdot)$) that computes the importance weights assigned to each candidate unit, and a router ($\mathcal{F}^\mathrm{R}(\cdot)$) that estimates the probability of each unit being selected.
Given the input feature $\bar{\mathit{f}}_{a}$, the output of the $a$-th dynamic selection module can be calculated by
\begin{equation}
   \mathit{\bar{M}}_{a} = \sum_{k\in\mathcal{T}} \mathcal{F}^\mathrm{W}(\bar{\mathit{f}}_{a})_k \cdot \mathcal{F}^\mathrm{U}_k(\bar{\mathit{f}}_{a}), 
\end{equation}
where $\mathcal{T} = \mathrm{TopK}(\mathcal{F}^\mathrm{R}(\bar{\mathit{f}}_{a}))$ denotes the set of selected top-$K$ candidate unit indices, and $\mathcal{F}^\mathrm{W}(\bar{\mathit{f}}_{a})_k$ is the importance weight for the $k$-th unit.

\begin{figure}[t] 
    \includegraphics[width=\linewidth]{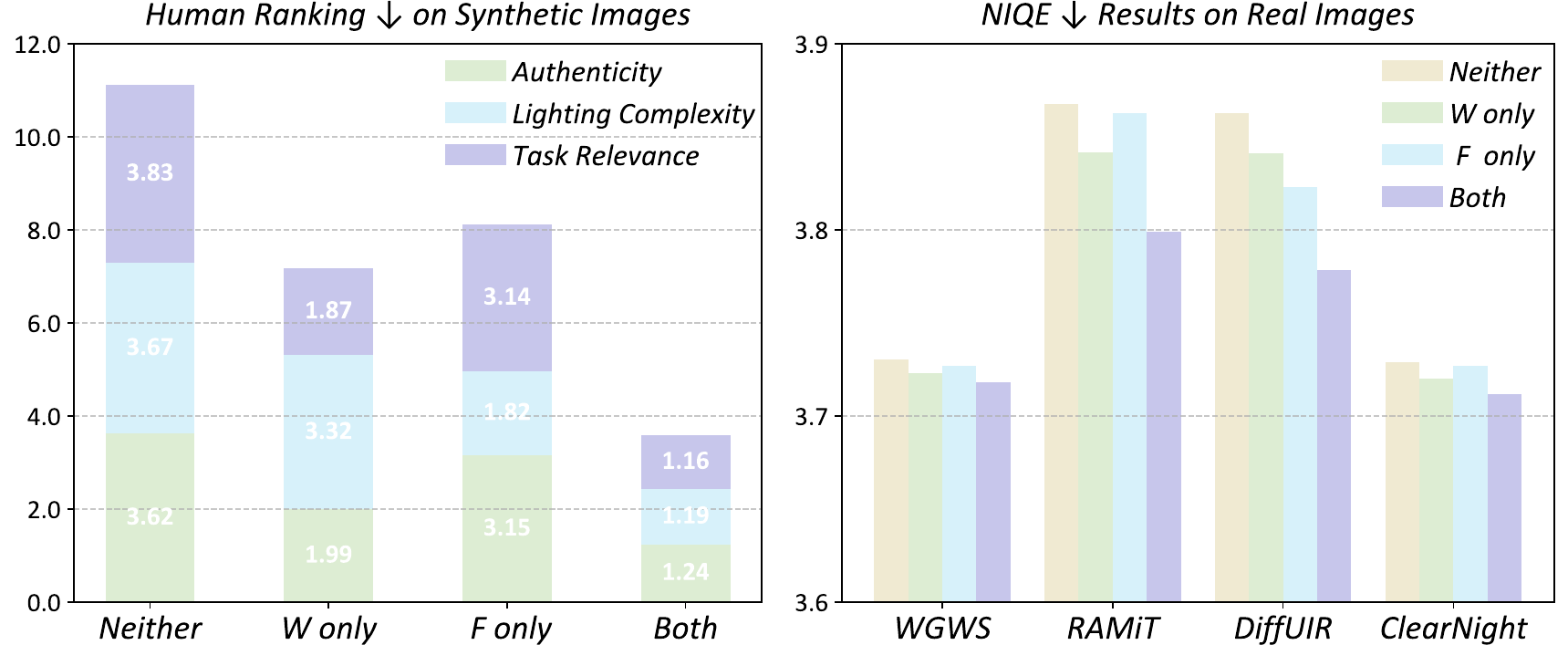}       
    \caption{Comparison of 4 synthesized variants of nighttime images, where \textbf{W} and \textbf{F} indicate synthesis with illumination-aware weather degradation and flare effects respectively.}
    \label{four_datasets} 
\end{figure}

\noindent\textbf{Dynamic Weather Degradation Modeling.} 
Relying solely on visual content would limit the ability to capture the correlations and distinctive characteristics of different weather types, thereby hindering the module's effectiveness in complex multi-weather scenarios.
To better associate distinct weather types with designated candidate units,  we reorganize the dynamic selection module and introduce a new component, the weather instructor ($\mathcal{F}^\mathrm{WI}(\cdot)$).

In the new design, features first pass through a knowledge-sharing unit ($\mathcal{F}^\mathrm{KSU}(\cdot)$), which is composed of shared linear layers preceding the dynamic allocator. The output of the knowledge-sharing unit is then utilized by the weather instructor, which classifies degradations and learns weather-specific prototypes to aggregate features from the same weather conditions:
\begin{equation}
\mathit{y}_{a}=\mathcal{F}^\mathrm{WI}(\bar{\mathit{f}}_a) \quad \textrm{with} \quad \bar{\mathit{f}}_a=\mathcal{F}^\mathrm{KSU}(\mathit{f}^\prime_a),
\end{equation}
where $\mathit{f}^\prime_a$ is the output of the encoder, and $\mathit{y}_a$ is the weather type predicted by the $a$-th dynamic selection module.

The weather instructor performs multi-label classification using binary cross-entropy loss $\mathcal{L}_\mathrm{bce}$, guiding the prototypes to attract features of the corresponding weather. 

\begin{figure}[t]
  \centering
  \includegraphics[width=1\linewidth]{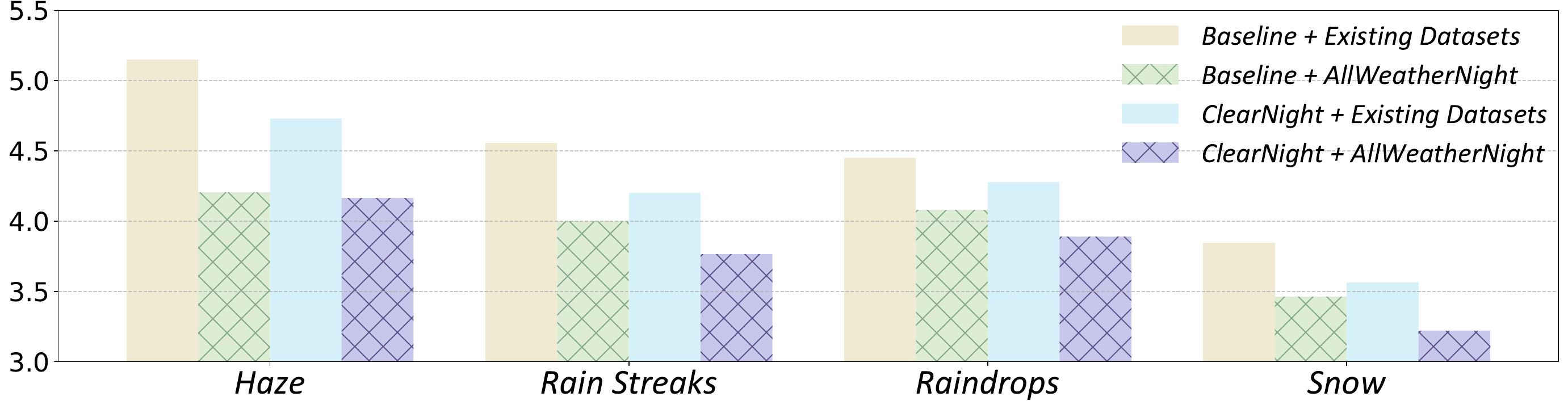}
   \caption{Models trained on AllWeatherNight achieve superior NIQE$\downarrow$ results on real-world images compared to those trained on the combination of existing nighttime datasets.}
   \label{otherdata}
\end{figure}

\begin{figure*}[t]
  \centering
  \includegraphics[width=1\linewidth]{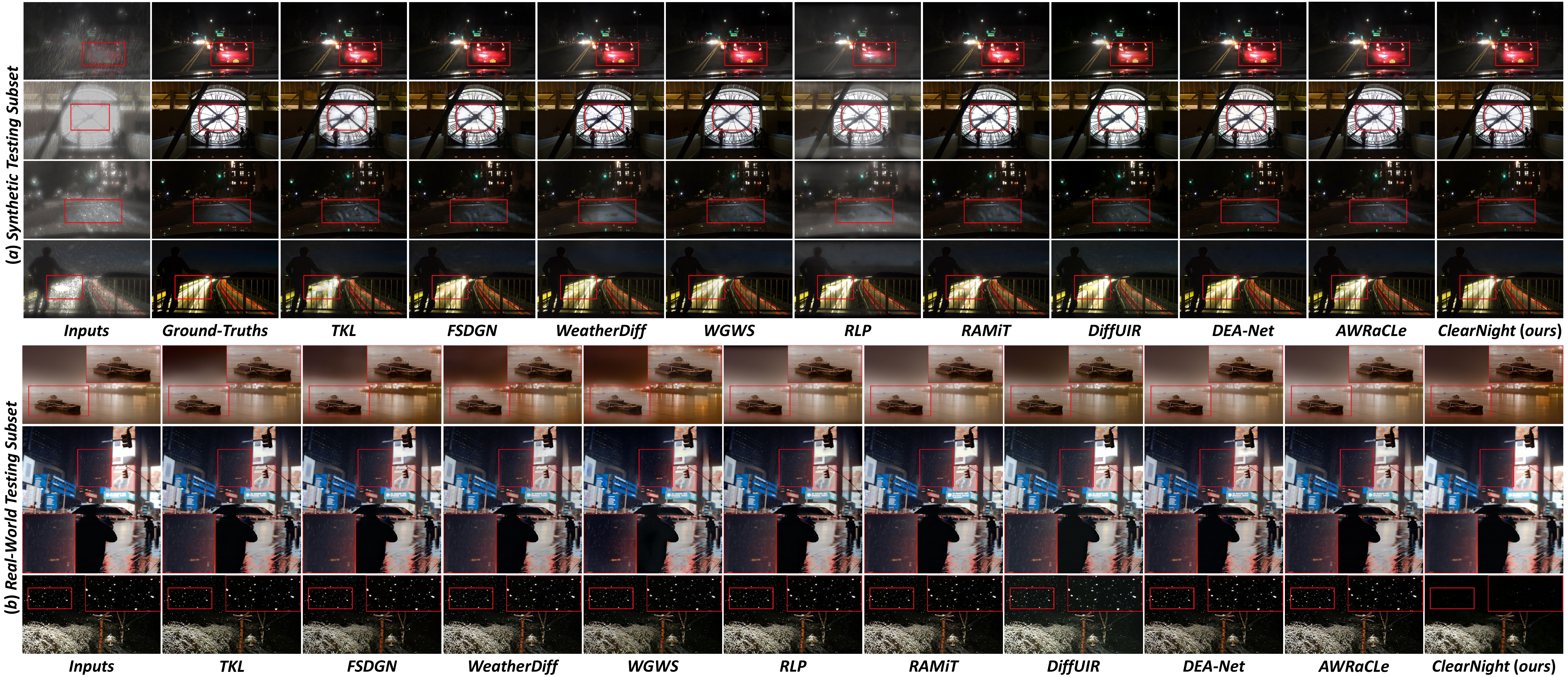}
   \caption{Qualitative results on AllWeatherNight synthetic and real-world testing subset. Please zoom in for more details.}
   \label{results_syn}
\end{figure*}

\begin{table*}[t!]
\small
\centering
\setlength{\tabcolsep}{0.5pt}
\setlength{\fboxsep}{0.8pt} 
\setlength{\fboxrule}{0.5pt} 
\begin{tabular}{l|cc|ccccc}
\hline
\rowcolor{mygray}~ & \multicolumn{2}{c|}{\textbf{Multi-Degradation}} &\multicolumn{5}{c}{\textbf{Single-Degradation}}\\
\rowcolor{mygray}\textbf{Method} & \textbf{Rain Scene} & \textbf{Snow Scene} & \textbf{Haze} &\textbf{Rain Streak} &\textbf{Raindrop}&\textbf{Snow}&  \textbf{Flare} \\
\rowcolor{mygray}~ & PSNR$\uparrow$ / SSIM$\uparrow$  & PSNR$\uparrow$ / SSIM$\uparrow$  & PSNR$\uparrow$ / SSIM$\uparrow$ & PSNR$\uparrow$ / SSIM$\uparrow$ &PSNR$\uparrow$ / SSIM$\uparrow$&PSNR$\uparrow$ / SSIM$\uparrow$ & PSNR$\uparrow$ / SSIM$\uparrow$ \\
\hline
TKL & 29.0919 / 0.8769 & 26.2657 / 0.8456 & 31.6136 / 0.9401& 30.4227 / 0.8844 & 32.5573 / 0.9561 & 31.0471 / 0.9247 & 36.7239 / 0.9741 \\
FSDGN  & 29.9850 / 0.8730 & 28.0147 / 0.8474 & 34.1807 / 0.9378 & 31.3309 / 0.8851 & 33.9535 / 0.9638 & 31.9022 / 0.9288 & 38.7223 / 0.9798\\
WeatherDiff & 29.7631 / 0.8936 & 27.1109 / 0.8537 & 30.5020 / 0.9256 & \fbox{33.3630} / \textbf{0.9352} & \textbf{35.9385} / \textbf{0.9726} & 33.6573 / 0.9488 & \fbox{40.2057} / 0.9826 \\
WGWS  & 30.5811 / 0.8961 & 27.9023 / 0.8658 & 31.6132 / 0.9181 & 32.8077 / 0.9221 & 34.3616 / 0.9554 & 32.1325 / 0.9196 & 32.1085 / 0.9017\\
RLP  & 21.2180 / 0.6641 & 19.3059 / 0.6124 & 18.6348 / 0.6379 & 31.1586 / 0.8867 & 32.4204 / 0.9322 & 31.5645 / 0.9101 & 34.7658 / 0.9451\\
RAMiT  & 30.4565 / 0.9106 & 29.1169 / 0.8889 & \fbox{36.4414} / \textbf{0.9738} & 31.9433 / 0.9204 & 33.8452 / 0.9632 & 32.8934 / 0.9491 & \textbf{43.0080} / \textbf{0.9934}\\
DiffUIR & 27.6676 / 0.8040 & 25.8151 / 0.7892 & 28.4763 / 0.8341 & 30.2547 / 0.8624 & 29.6778 / 0.8740 & 26.9833 / 0.8055 & 31.2789 / 0.8816\\
DEA-Net & 31.4244 / 0.9202 & 29.2500 / \fbox{0.8956} & 35.8174 / 0.9612 & 32.7631 / 0.9285 & 34.8406 / 0.9704 & 33.5811 / 0.9493 & 38.6533 / 0.9807 \\ 
AWRaCLe & \fbox{31.5392} / \fbox{0.9210} & \fbox{29.4270} / 0.8738 & 36.4315 / 0.9599 & 33.1078 / 0.9317 & 35.2985 / 0.9686 & \fbox{33.6716} / \fbox{0.9532} & 40.1014 / 0.9836 \\ 
\hline
Baseline & 28.7976 / 0.8825 & 27.1337 / 0.8452 & 30.2905 / 0.9257 & 30.5615 / 0.8994 & 33.0758 / 0.9598 & 31.3318 / 0.9182 & 36.0821 / 0.9738\\
ClearNight
& \textbf{32.5937} / \textbf{0.9223} & \textbf{30.6464} / \textbf{0.9041} & \textbf{36.4655} / \fbox{0.9621} & \textbf{33.6238} / \fbox{0.9331} & \fbox{35.4282} / \fbox{0.9723} & \textbf{33.9747} / \textbf{0.9539} & 38.7707 / \fbox{0.9838}\\
\hline
\end{tabular}
\caption{Quantitative results on AllWeatherNight synthetic testing subset. The \textbf{best} and \fbox{second-best} results are highlighted.}
\label{syn}
\end{table*}

\subsection{Network Optimization}
Multiple losses are utilized to jointly optimize ClearNight, where L1 loss $\mathcal{L}_\mathrm{1}$ and perceptual loss $\mathcal{L}_\mathrm{p}$~\cite{per} are used to ensure that restored results resemble ground-truths closely. 
To enhance the structure and details of outputs, depth loss $\mathcal{L}_\mathrm{d}$ is exploited via minimizing the differences between ground-truths and predictions of a pre-trained depth estimation model~\cite{depth}.
In addition, we use a load balancing loss $\mathcal{L}_\mathrm{lb}$~\cite{losslb} to balance the utilization of the candidate units. 
We jointly optimize the network using the total loss function:
\begin{equation}
   \mathcal{L}_\mathit{total} = \mathcal{L}_\mathrm{1} + \lambda_\mathrm{p} \mathcal{L}_\mathrm{p} + \lambda_\mathrm{bce} \mathcal{L}_\mathrm{bce} + \lambda_\mathrm{lb} \mathcal{L}_\mathrm{lb} + \lambda_\mathrm{d} \mathcal{L}_\mathrm{d},
\end{equation}
where $\lambda_\mathrm{p}$, $\lambda_\mathrm{bce}$, $\lambda_\mathrm{lb}$ and $\lambda_\mathrm{d}$ are the loss weights.

\section{Experiments}
\label{sec:experiment}

\subsection{Implementation Details}
The synthetic image size is 640 $\times$ 360. During training, the input image is randomly cropped to 256 $\times$ 256. 
We adopt DehazeFormer~\cite{swin} as our baseline for its advanced spatial aggregation, allowing effective natural lighting restoration.
We use Adam as optimizer and the initial learning rate is set to $2 \times 10^{-4}$ for 100 epochs. The learning rate is adjusted using the cosine annealing scheme. The loss weight $\lambda_\mathrm{p}$, $\lambda_\mathrm{bce}$, $\lambda_\mathrm{lb}$ and $\lambda_\mathrm{d}$ are empirically set to 0.1, 0.001, 0.01 and 0.02, respectively.
$\alpha$ is set to 0.995.

\subsection{Dataset Analysis}
To demonstrate the effectiveness of illumination-aware degradation generation, we analyze four different synthetic variants of nighttime images (showcased in Fig.~\ref{visual_dataset}).
Ten volunteers are recruited to rank image quality on authenticity, lighting complexity and task relevance.
As shown in the left of Fig.~\ref{four_datasets}, we can observe that the combination of illumination-aware weather and flare achieves the best human preference.
Furthermore, the right part of Fig.~\ref{four_datasets} demonstrates that training with our illumination-aware data improves performance in real-world scenes, with four representation models~\cite{LWGWSCVPR2023,RAMITCVPR2024,SCVPR2024} exhibiting consistent improvements.

To further validate the proposed dataset, we compare the performance of models trained on our AllWeatherNight dataset with those trained on a composite dataset constructed from existing nighttime datasets, as shown in Tab.~\ref{datasets}.
As shown in Fig.~\ref{otherdata}, models trained on AllWeatherNight exhibit superior performance on real-world samples, which is attributable to our illumination-aware synthetic images simulating degradations more realistically.

\subsection{Comparison with the State-of-the-Art}
\noindent\textbf{Results on Synthetic Data.} 
We compare ClearNight with state-of-the-art adverse weather image restoration methods on AllWeatherNight, including TKL~\cite{TKLCVPR2022}, FSDGN~\cite{FSDGNECCV2022}, RLP~\cite{NICCV2023}, WeatherDiff~\cite{WeatherDiff2023}, WGWS~\cite{LWGWSCVPR2023}, RAMiT~\cite{RAMITCVPR2024}, DiffUIR~\cite{SCVPR2024}, DEA-Net~\cite{TIP2024DEA} and AWRaCLe~\cite{AAAAI2025}. 
As shown in Fig.~\ref{results_syn}a, TKL~\cite{TKLCVPR2022}, FSDGN~\cite{FSDGNECCV2022} and DiffUIR~\cite{SCVPR2024} produce overly smooth results such as the clock and railing regions, while other approaches often lose structural details or leave residual artifacts.
In contrast, ClearNight preserves rich background details and restores natural lighting, illustrating robust performance across diverse and complex nighttime scenarios.

Tab.~\ref{syn} reports quantitative results on multi-degradation and single-degradation scenes, evaluated using PSNR~\cite{PSNR} and SSIM~\cite{SSIM} for synthetic samples. 
Among these methods, FSDGN, RLP and DEA-Net are tailored for daytime/nighttime task-specific restoration, but struggle in other task scenes due to limited weather feature extraction.  
In contrast, our method targets robust nighttime weather effects removal. As the training data lack dedicated flare samples, its performance in flare removal remains moderate. 
Nevertheless, ClearNight achieves the best results on the multi-degradation subset and delivers competitive performance on single-degradation samples.

\noindent\textbf{Results on Real-World Data.} 
As shown in Fig.~\ref{results_syn}b, ClearNight effectively removes most weather effects on real images and mitigates flares, producing more natural results compared to state-of-the-art methods.
Furthermore, we use NIQE~\cite{NIQE} to assess the restored images in Tab.~\ref{real}. The experimental results demonstrate that ClearNight can predict highly realistic images under various adverse weather conditions.

\begin{figure}[t]
\centering   
\includegraphics[width=0.98\linewidth]{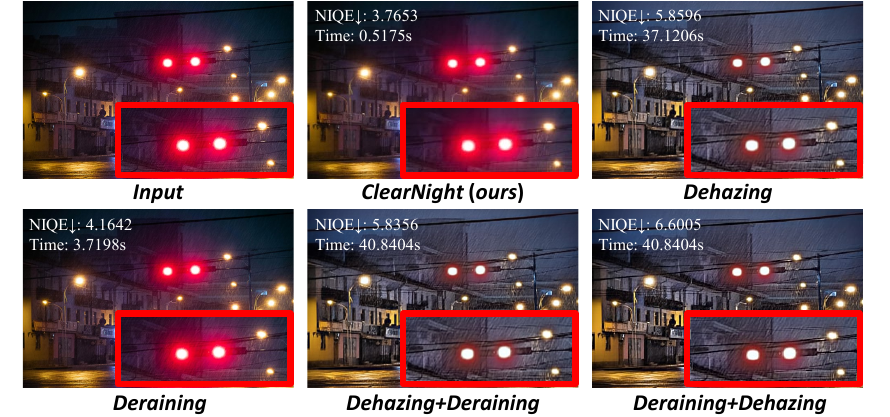}
\caption{ClearNight outperforms a cascade of two latest task-specific nighttime restoration methods. 
Quantitative results are averaged over all the real-world rain streak images.}
\label{mix_method}
\end{figure}

\noindent\textbf{Comparison with Cascade Solutions.} 
As demonstrated in Fig.~\ref{mix_method}, we compare ClearNight against two pre-trained task-specific models~\cite{NAAAI2025,NICCV2023} on real-world nighttime rain streak images from AllWeatherNight. 
ClearNight not only achieves superior results visually and quantitatively, but requires significantly less inference time.

\begin{table}[t]
\small
\centering
\setlength{\tabcolsep}{5pt}
\setlength{\fboxsep}{1pt} 
\setlength{\fboxrule}{0.5pt} 
\begin{tabular}{l|cccc}
\hline
\rowcolor{mygray}\textbf{Method} & \textbf{Haze} &\textbf{Rain Streak} & \textbf{Raindrop} & \textbf{Snow} \\
\hline
TKL & 4.1872 & 3.7765 & \fbox{3.9238} & \fbox{3.2680} \\
FSDGN  & 4.2780 & 4.4694 & 4.9149 & 4.1528 \\
WeatherDiff & 4.1964 & 3.7842 & 3.9254 & 3.3451 \\
WGWS  & 4.1879 & \fbox{3.7732} & 3.9635 & 3.2769 \\
RLP & 4.9699  & 4.1882  & 5.6240  & 4.7669 \\
RAMiT  & 4.1655 & 3.9298 & 4.0808  & 3.2497  \\
DiffUIR  & \textbf{4.1063}  & 3.8728  & 4.0471  & 3.3547 \\
DEA-Net  & 4.1665 & 3.9826 & 4.0889 & 3.4201 \\
AWRaCLe  & 4.1681 & 3.9516 & 4.0936 & 3.3398 \\
\hline
Baseline & 4.2054 & 3.9983 & 4.0778 & 3.4605\\
ClearNight
& \fbox{4.1623} & \textbf{3.7653} & \textbf{3.8882} & \textbf{3.2191} \\
\hline
\end{tabular}
\caption{Quantitative results on AllWeatherNight real-world testing subset, evaluated using the commonly used NIQE$\downarrow$.}
\label{real}
\end{table}

\begin{table}[t!]
\small
\centering
\setlength{\tabcolsep}{10.5pt}
\begin{tabular}{c|cccc|c}
\hline
\rowcolor{mygray}\textbf{\#} & \textbf{IP} & \textbf{DA} & \textbf{RP} & \textbf{WI} & \textbf{PSNR$\uparrow$ / SSIM$\uparrow$}\\
\hline
1 & \ding{56} & \ding{56} & \ding{56} & \ding{56} & 28.7976 / 0.8825\\
2 & \ding{52} & \ding{56} & \ding{56} & \ding{56} & 32.1304 / 0.9176\\
3 & \ding{56} & \ding{52} & \ding{56} & \ding{56} & 31.7075 / 0.9113\\
4 & \ding{52} & \ding{52} & \ding{56} & \ding{56} & 32.2393 / 0.9179\\
5 & \ding{56} & \ding{52} & \ding{52} & \ding{52} & 32.4430 / 0.9214\\
6 & \ding{52} & \ding{52} & \ding{52} & \ding{56} & 32.2528 / 0.9184\\
7 & \ding{52} & \ding{52} & \ding{52} & \ding{52} & \textbf{32.5937} / \textbf{0.9223}\\
\hline
\end{tabular}
\caption{Ablation study of key component. \textbf{IP} and \textbf{RP} denote illumination and reflectance priors. \textbf{DA} and \textbf{WI} indicate dynamic allocator and weather instructor in WDS.}
\label{model}
\end{table}

\subsection{Ablation Studies}

\noindent\textbf{Ablation on Model Components.} 
We evaluate the effectiveness of the illumination prior, reflectance prior, dynamic allocator and weather instructor on the Rain Scene testing subset.
As shown in Tab.~\ref{model}, dual priors significantly enhance the restoration performance over the baseline, 
while the dynamic allocator, guided by the weather instructor, optimizes candidate unit selection to handle multi-weather scenes. 
Quantitative results demonstrate that the integration of all components achieves the best performance. 

\begin{figure}[t]
\centering
\includegraphics[width=1\linewidth]{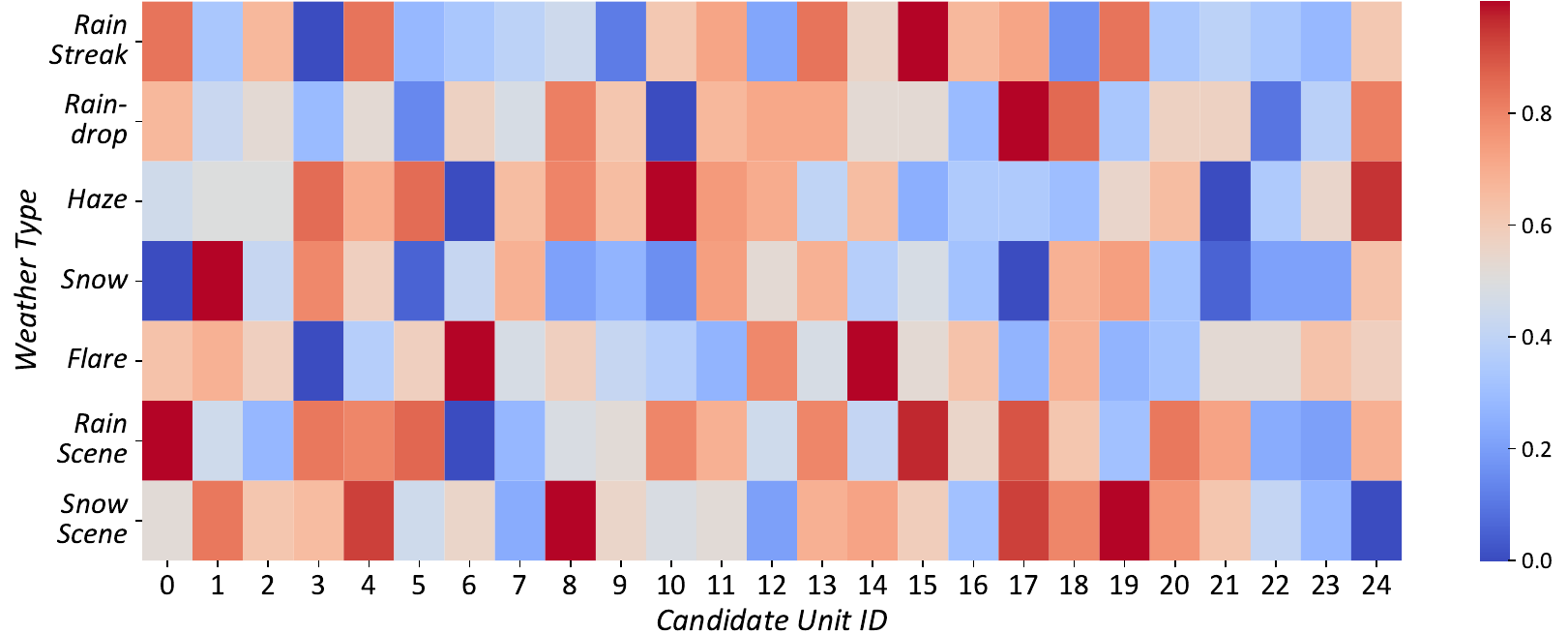}
\caption{Correlation of selected units and diverse degradations. WDS associates various distortions with candidate units.}
\label{corr}
\end{figure}

\begin{figure}[t!]
   \includegraphics[width=\linewidth]{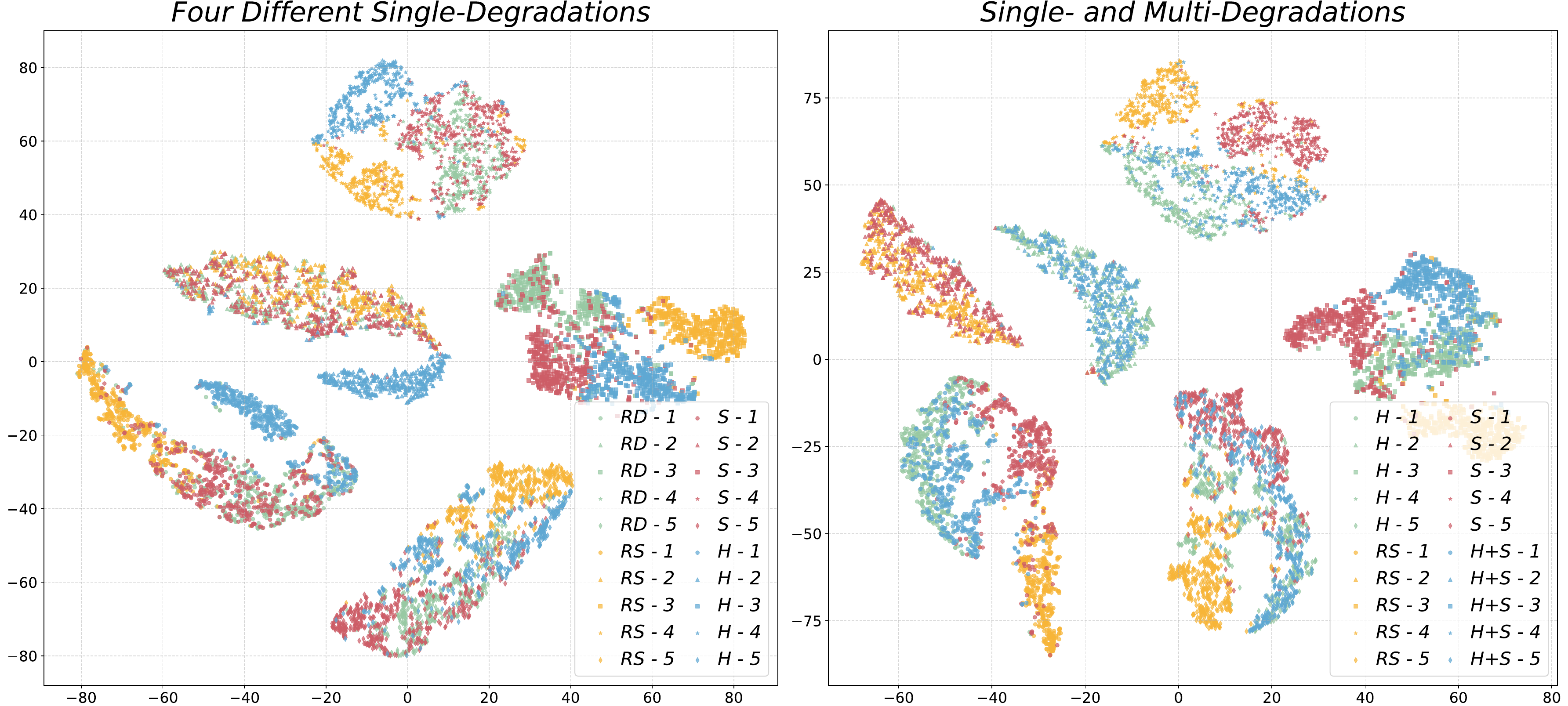}
   \caption{T-SNE visualization of feature distributions across distinct weather. \textbf{H}, \textbf{RS}, \textbf{RD} and \textbf{S} indicate haze, rain streak, raindrop and snow. The number is the index of WDS.}
   \label{t-SNE2}
\end{figure}

\noindent\textbf{Effectiveness of WDS.} 
Fig.~\ref{corr} illustrates the relationships between various degradations and selected units, demonstrating the effectiveness of WDS in capturing the specificity of different degradations. 
The WDS differentiates weather features, selecting similar unit combinations for degraded scenes with the same weather, enabling the network to efficiently eliminate diverse distortions. 
We visualize the feature $\bar{\mathit{f}}^a$ of different weather types in Fig.~\ref{t-SNE2}. Despite the visual similarity between rain and snow, ClearNight still differentiates them. Notably, the features of ``$\mathrm{H+S}$'' lie between haze and snow features, indicating that our model learns the correlations among multiple weather effects in complex nighttime scenarios.

\section{Conclusion}\label{sec:conclusion}
This paper explores a practical yet under-explored task, \textit{i.e.}, multi-weather nighttime image restoration. 
To facilitate this task, we introduce an illumination-aware degradation generation approach and construct a new dataset featuring 10K high-quality nighttime images with various compositional adverse weather and lighting conditions.
In addition, we propose ClearNight, a unified framework, tailored for the new task, capable of removing multiple degradations in one go.
ClearNight leverages Retinex-based dual priors to explicitly guide the network to focus on illumination regions and intrinsic textures respectively.
Moreover, ClearNight incorporates weather-aware dynamic specific-commonality collaboration to better capture the characteristics of diverse weather, enabling effective multi-weather degradation removal.
Comprehensive experiments on both synthetic and real-world images demonstrate the superiority of our ClearNight.

\begin{figure*}[t]
\begin{minipage}{1\textwidth} 
    \centering
    \Large\bfseries 
    Clear Nights Ahead: Towards Multi-Weather Nighttime Image Restoration\\
    Supplementary Material \\ 
\end{minipage}
\end{figure*}

\section{Additional Implementation Details}\label{sec: additional imp}

As detailed in the main text, the model was trained on 256 $\times$ 256 randomly cropped patches. For inference, full-resolution images were used.
The model was trained with a batch size of 1. 
To stabilize training, we adopt a cosine annealing warm restarts strategy with a restart period of 50 epochs. 
We use the Adam optimizer~\cite{Adam} with $\beta_1$=0.9 and $\beta_2$=0.99.
The proposed framework is implemented in Pytorch and was trained and tested on an NVIDIA RTX A6000. 

The architecture of ClearNight, applies a multi-branch structure \cite{NICCV2023,UCVPR2023,LWGWSCVPR2023}, integrating illumination and reflectance priors into the backbone. 
The backbone consists of a commonality branch and a weather-aware dynamic specificity branch. 
The commonality branch employs five-stage Transformer blocks \cite{swin} with modified normalization layers, activation functions, and spatial feature aggregation to enhance global feature extraction for improved restoration performance.
The weather-aware dynamic specificity branch includes five weather-aware dynamic selection blocks (WDS) tailored to adverse weather degradations. After each interaction between unique and common features, we use upsampling and downsampling operations to adjust feature map scales, doubling or halving the width and height as needed.

\section{General Discussions}
\subsection{Dataset License and Intended Use}\label{sec: discussion}
The generated images and labels in AllWeatherNight dataset are released under the BSD 3-Clause License, a permissive open-source license that grants users the freedom to use, copy, modify, and distribute the dataset, whether in its original form or as part of derivative works. The ground-truths are released under the BSD 3-Clause License.

The AllWeatherNight dataset is designed to advance research in multi-weather nighttime image restoration. It provides a comprehensive resource for developing, training, and evaluating algorithms to restore nighttime images degraded by diverse adverse weather degradations and non-uniform flare effects. Furthermore, it serves as a standardized benchmark for comparing methods in adverse weather image restoration field.

\subsection{Limitations and Future Works}\label{sec: limitations}

\begin{figure}[t]
  \centering
  \includegraphics[width=\linewidth]{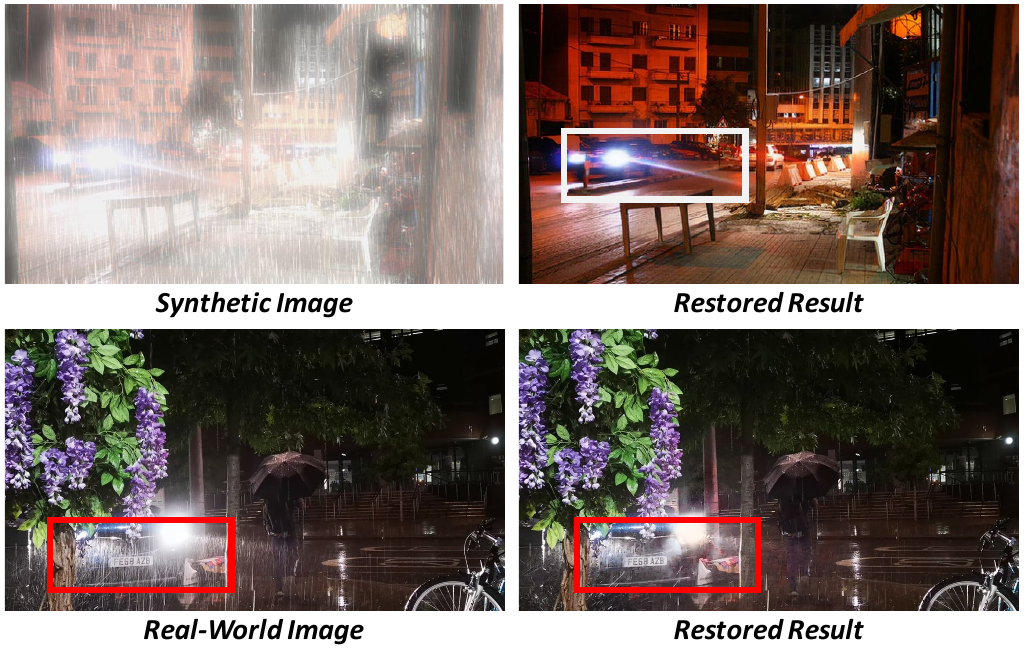}
   \caption{Adverse weather samples and their restored results under rapidly changing lighting conditions.}
   \label{limitation}
\end{figure}

\begin{figure*}[t]
    \includegraphics[width=\textwidth]{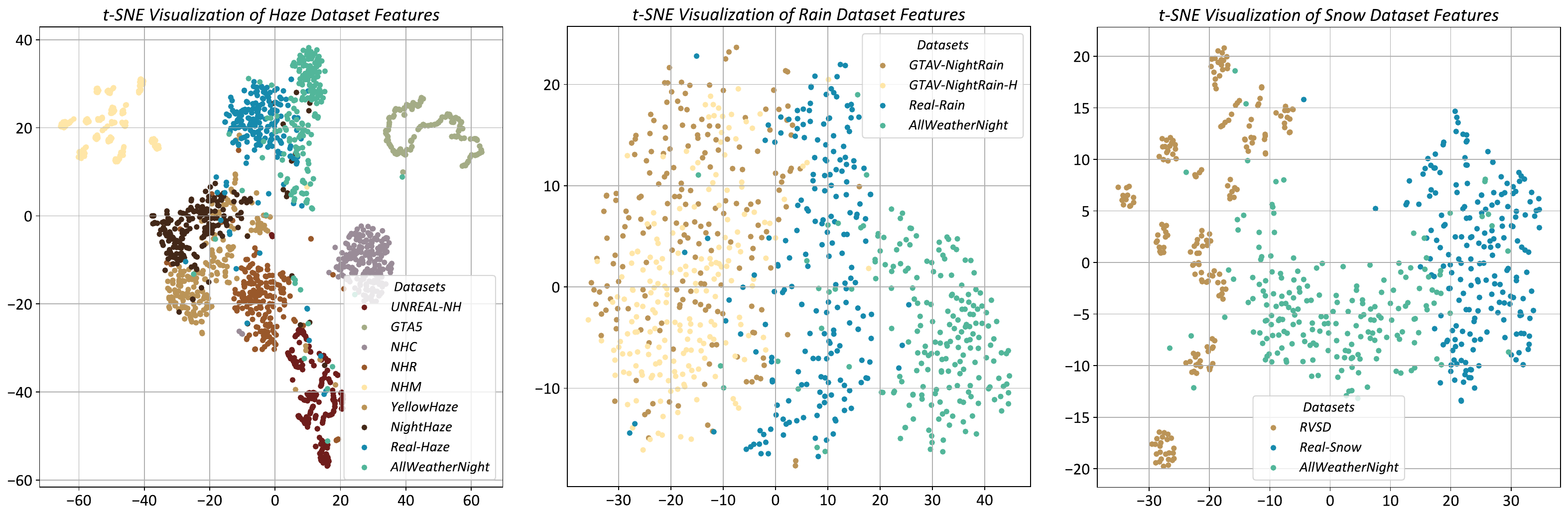}
    \caption{The t-SNE map of various nighttime adverse weather datasets, including haze, rain and snow degradations. The feature distribution of adverse weather degradations in our AllWeatherNight is closer to real-world nighttime scenes than existing nighttime synthetic datasets.}
    \label{t-sne}
\end{figure*}

\begin{figure*}[t]
    \includegraphics[width=\textwidth]{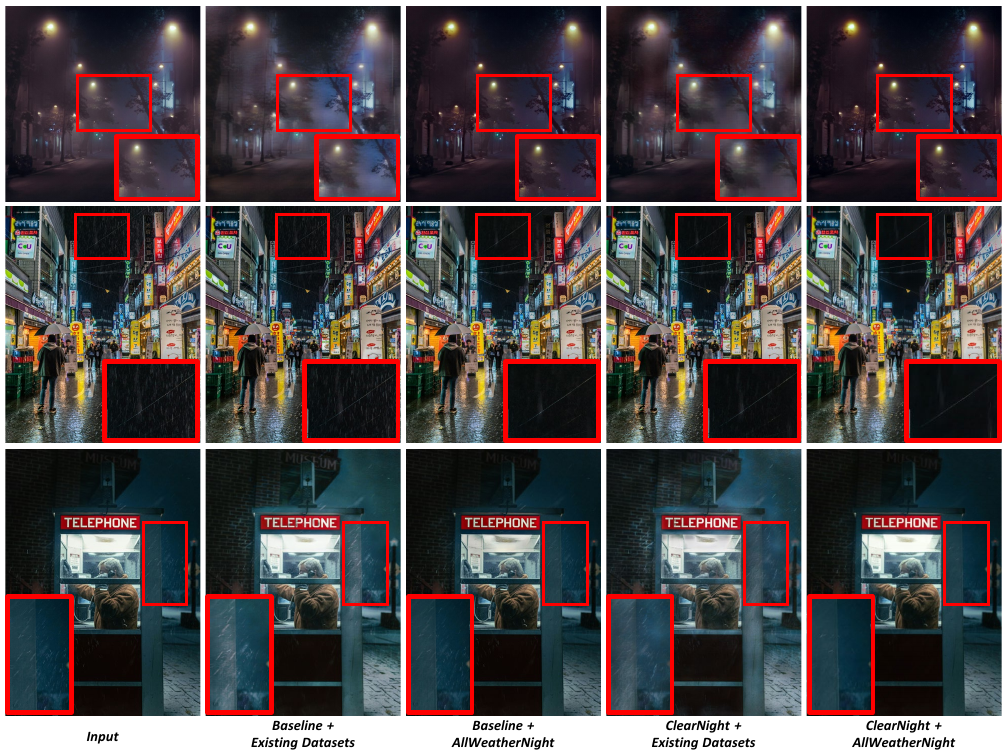}
    \caption{Models trained with AllWeatherNight achieve superior performance on real-world images compared to those trained with the combination of existing nighttime adverse weather datasets in Tab.~1.}
    \label{otherdata_ourdata}
\end{figure*}

While our approach effectively handles nighttime illumination and various adverse weather degradations, its performance under extremely dynamic lighting conditions, such as rapidly changing artificial lighting, needs improvement. As shown in Fig.~\ref{limitation}, while most weather-induced degradations can be removed, rapidly changing flares still remain. Future work could focus on improving the robustness to such extreme scenarios. The original data for AllWeatherNight are sourced from BDD100K \cite{BDD100K} and Exdark \cite{Exdark}, which exclusively focus on driving and object detection. Although our dataset incorporates various nighttime adverse weather degradations, the natural scenes in it are relatively homogeneous, limiting its environmental diversity. We consider expanding the AllWeatherNight to include a wider range of real-world scenarios to further enhance the model's generalization ability.

\section{Further Analysis of AllWeatherNight Dataset}\label{sec: datasets}

\subsection{Authenticity Assessment} 
To validate the authenticity of the AllWeatherNight dataset, we conduct a statistical analysis of feature distribution, comparing it against existing synthetic adverse weather nighttime datasets and real-world samples. 
The nighttime synthetic datasets include various haze datasets (UNREAL-NH~\cite{NACM20232}, GTA5~\cite{NECCV2020}, NHC/NHR/NHM~\cite{NACM2020}, NightHaze/YellowHaze~\cite{NPCM2018}), rain datasets (GTAV-NightRain~\cite{GTAV}, GTAV-NightRain-H~\cite{NICCV2023}) and snow dataset (RVSD~\cite{SICCV2023}). 
As shown in Fig.~\ref{t-sne}, the t-SNE visualization reveals that AllWeatherNight's adverse weather degradation distributions are more closely aligned with real-world nighttime images than any other synthetic nighttime dataset. 
From left to right, the subfigures depict the feature distributions for haze, rain, and snow, respectively. 
Notably, AllWeatherNight’s haze distribution closely matches that of real-world nighttime haze, while its rain and snow distributions exhibit the highest similarity to real-world adverse weather samples.
These findings demonstrate that AllWeatherNight outperforms existing synthetic datasets in approximating real-world nighttime degradation features, making it highly suitable for training models that generalize to diverse real-world scenarios.
Consequently, AllWeatherNight provides a robust foundation for enhancing model performance in multi-weather nighttime image restoration tasks.

\subsection{Effectiveness on Real-World Data}
To validate the effectiveness of our AllWeatherNight, we train baseline and ClearNight models on our dataset and on a composite dataset, which includes all existing nighttime single weather datasets listed in Tab.~1 of the main text.
As shown in Fig.~\ref{otherdata_ourdata}, both the baseline and our ClearNight models trained on AllWeatherNight consistently produce high-quality results on real-world nighttime scenarios. 
In contrast, models trained on the composite dataset tend to generate artifacts and often leave adverse weather degradations in the restored results. 
This underperformance can be attributed to the composite dataset's fundamental failure to account for the co-occurrence of multiple weather conditions and the complex entanglement of these degradations with uneven lighting, consequently lacking real-world fidelity and realism.
Consistent with quantitative results in Fig. 7 of the main text, these visual results demonstrate the superior quality and effectiveness of our AllWeatherNight for multi-weather nighttime image restoration. 

\begin{figure*}[t]
  \centering
  \includegraphics[width=1\linewidth]{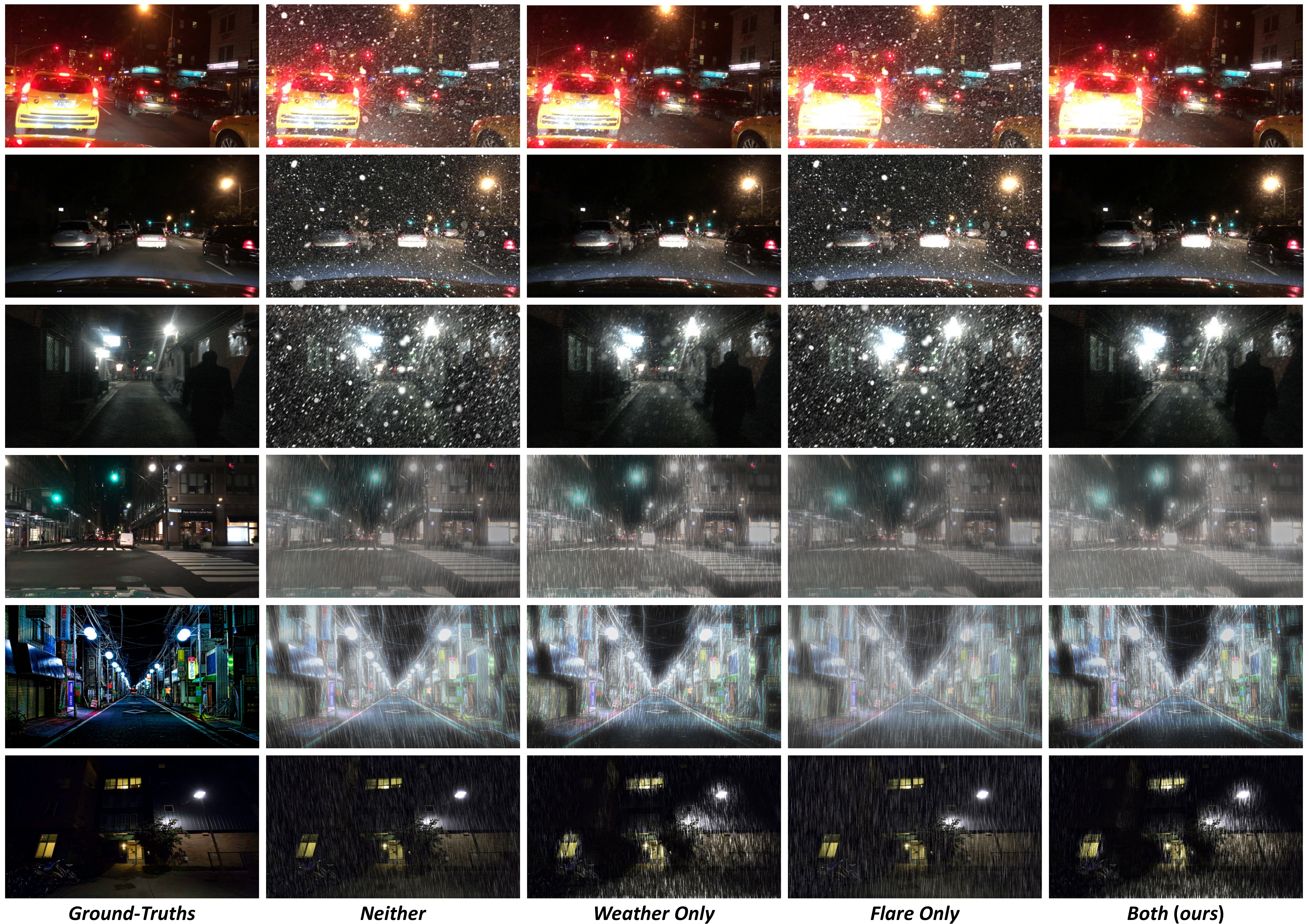}
  \caption{Visualization of 4 different synthesized variants of adverse weather nighttime images, where \textbf{Weather only} and \textbf{Flare only} indicate synthesis with illumination-aware weather degradation and flare degradation respectively. Ours includes both degradation synthesis.}
  \label{datasets2}
\end{figure*}

\begin{figure*}[h]
  \centering
  \includegraphics[width=1\linewidth]{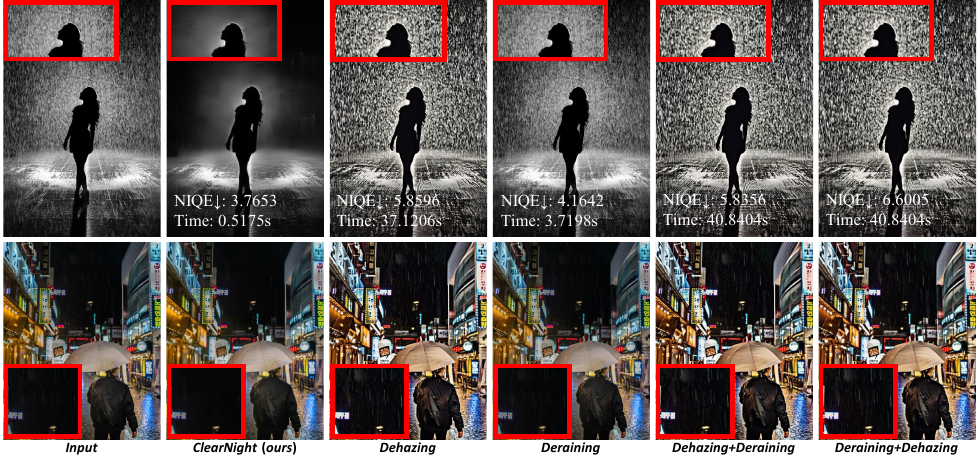}
   \caption{ClearNight outperforms a cascade of two latest task-specific nighttime adverse weather image restoration methods. Quantitative results are averaged over all the real-world rain streak images.}
   \label{mix_method2}
\end{figure*}

\begin{figure*}[t!]
    \includegraphics[width=\textwidth]{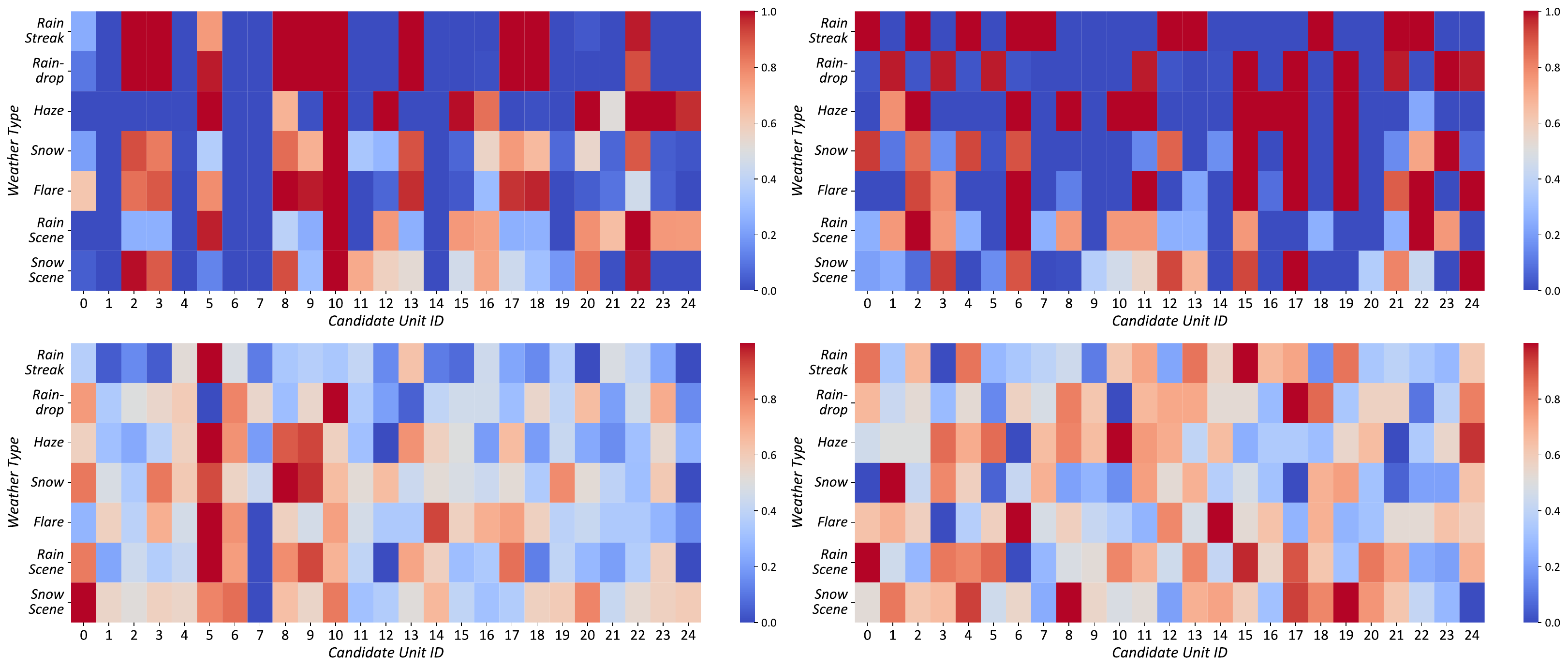}
    \caption{Correlation between degradation types and candidate units correlation under different loss supervision. Top left: without$ \mathcal{L}_\mathrm{b}$ and $\mathcal{L}_\mathrm{lb}$. Top right: with only $\mathcal{L}_\mathrm{b}$. Bottom left: with only $\mathcal{L}_\mathrm{lb}$. Bottom right: with both $\mathcal{L}_\mathrm{b}$ and $\mathcal{L}_\mathrm{lb}$.}
    \label{two losses}
\end{figure*}

\begin{figure}[h]
  \centering
  \includegraphics[width=1\linewidth]{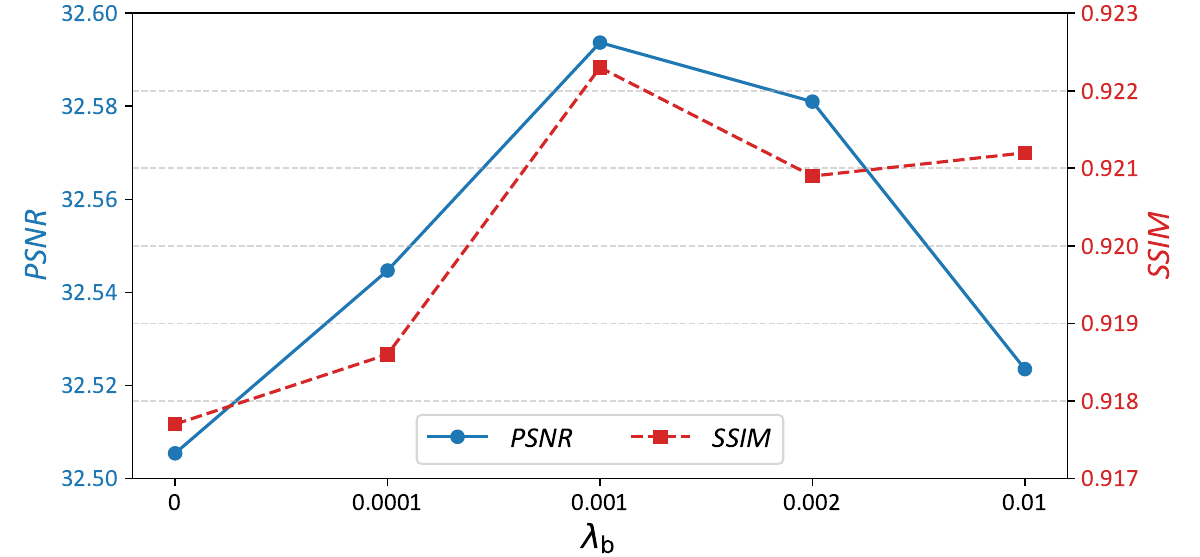}
   \caption{Ablation study on $\lambda_\mathrm{b}$.}
   \label{bce}
\end{figure}

\subsection{Additional Visualizations of Illumination-Aware Degradation Generation} 
Existing synthetic adverse weather nighttime datasets typically model a single type of weather distortion and rarely account for background blurring induced by artificial lighting. 
In contrast, the proposed AllWeatherNight encompasses a diverse range of degradation, including haze, rain streaks, raindrops, snow, flare, and their combinations. 
Moreover, we design an illumination-aware degradation generation method to simulate realistic adverse weather and flare effects in nighttime scenes.
We synthesized datasets under four settings, with supplemental visualizations provided in Fig.~\ref{datasets2}. 
Each type of degradation in AllWeatherNight varies in scales, directions, densities, and styles, ensuring comprehensive coverage. 
To simulate these degradations, we apply the atmospheric scattering model and interpolated Gaussian noise for haze and rain streaks~\cite{HCVPR2019}, utilize Bezier curves for raindrop synthesis~\cite{Raindrop}, and add snow degradation using snowflake masks from~\cite{snow100k,csd}.
The generated flare preserves background clarity in the original image to some extent, enabling the network to effectively restore natural lighting and fine details.

\section{Further Analysis of ClearNight}
\subsection{ClearNight \textit{vs.} Cascaded Solutions}

To further validate the necessity of our unified framework, ClearNight, for restoring complex nighttime images under multiple adverse weather conditions, we present comprehensive analysis comparing our unified model against two pre-trained task-specific models~\cite{NICCV2023,NAAAI2025} and their cascaded pipeline. 
For the cascaded pipeline, we evaluate two configurations: one that applies the dehazing model followed by the deraining model (dehazing + deraining) and the other in the reverse order (deraining + dehazing). 
As illustrated in Fig.~\ref{mix_method2}, task-specific models fail to remove multiple weather degradations, and their cascaded models struggle to produce satisfactory results. In contrast, our ClearNight generates more visually pleasing results and achieves the best average NIQE scores with superior computational efficiency. 
\subsection{Additional Ablation Studies}\label{sec: DWS}

\noindent\textbf{Influence of Retinex-Based Dual Prior Guidance.} 
To evaluate the impact of multi-scale dual prior, we conduct ablation studies on illumination and reflectance priors at various scales, using \# 1 and \# 3 (in Tab. 4 of the main text) as baselines, respectively. As shown in Tabs.~\ref{igp}-\ref{rdp}, incorporating multi-scale illumination and reflectance priors into the network enables it to focus on uneven lighting and intrinsic texture features of different scales, significantly enhancing nighttime image restoration performance. As the inclusion of illumination components across all five stages brings only marginal gains (PSNR + 0.1628, SSIM + 0.0022), we adopt them only in the first three stages to achieve a better trade-off between performance and model complexity.

\begin{figure}[t!]
\centering
\includegraphics[width=1\linewidth]{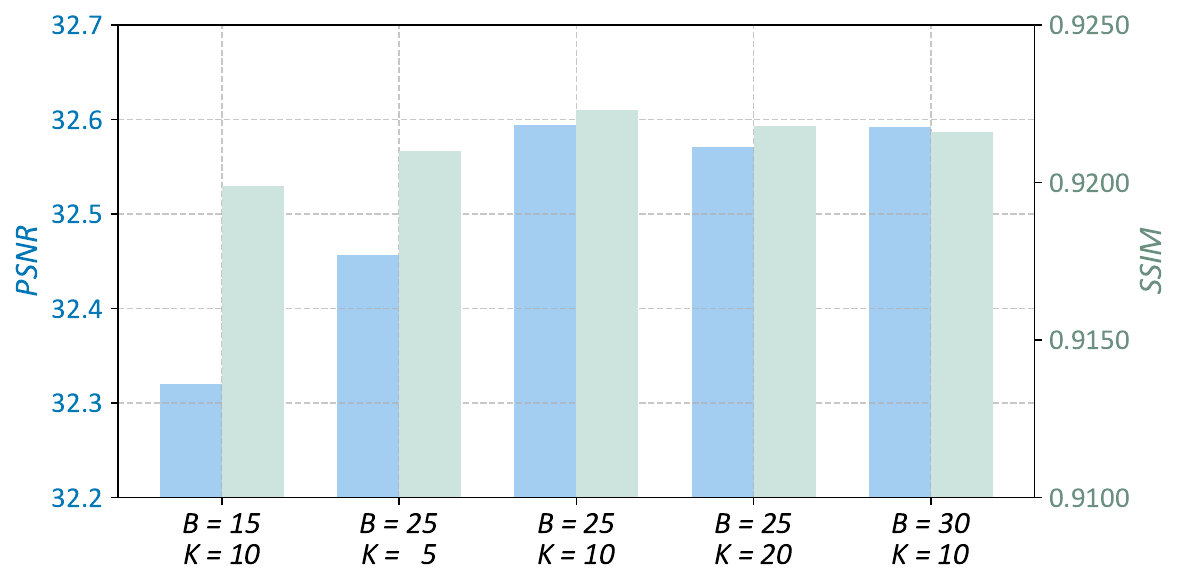}
\caption{Ablation study of the number of candidate units $B$, and selected top-$K$ units. We set $B$ and $K$ to 25 and 10 by default.}
\label{b-k}
\end{figure}

\noindent\textbf{Effectiveness of DSM.} 
We investigate the number of candidate units ($B$) and top-$K$ selection parameter ($K$). As shown in Fig.~\ref{b-k}, increasing the number of candidate units enhances selective capacity, with an optimal $K$ being critical for superior performance. Notably, performance improves with larger $K$ up to a threshold, beyond which irrelevant information causes model degradation. 

\begin{table}[t]
\caption{Ablation on Retinex-based Illumination Prior.}
        \centering
        \footnotesize
        \setlength{\tabcolsep}{14pt}
        \begin{tabular}{c|ccc|c}
        \hline
        \rowcolor{mygray} \textbf{\#} &$\mathit{i}_{1}$ & $\mathit{i}_2$ & $\mathit{i}_3$ & PSNR$\uparrow$ / SSIM$\uparrow$\\
        \hline
        1 & \ding{56} & \ding{56} & \ding{56} & 28.7976 / 0.8825\\
        2 & \ding{52} & \ding{56} & \ding{56} & 31.6843 / 0.9109 \\
        3 & \ding{56} & \ding{52} & \ding{56} & 31.6714 / 0.9114\\
        4 & \ding{56} & \ding{56} & \ding{52} & 31.6429 / 0.9110\\
        5 & \ding{52} & \ding{52} & \ding{56} & 31.7217 / 0.9115\\
        6 & \ding{52} & \ding{52} & \ding{52} & \textbf{32.1304} / \textbf{0.9176}\\
        \hline
        \end{tabular}
        \label{igp}
\end{table}

\begin{table}[t]
\caption{Ablation on Retinex-based Reflection Prior.}
        \centering
        \footnotesize
        \setlength{\tabcolsep}{14pt}
        \begin{tabular}{c|ccc|c}
        \hline
        \rowcolor{mygray} \textbf{\#}&$\mathit{r}_1$ & $\mathit{r}_2$ & $\mathit{r}_3$ & PSNR$\uparrow$ / SSIM$\uparrow$\\
        \hline
        1 & \ding{56} & \ding{56} & \ding{56} & 31.7075 / 0.9113 \\
        2 & \ding{52} & \ding{56} & \ding{56} & 31.9680 / 0.9198\\
        3 & \ding{56} & \ding{52} & \ding{56} & 31.9374 / 0.9201\\
        4 & \ding{56} & \ding{56} & \ding{52} & 31.8438 / 0.9172\\
        5 & \ding{52} & \ding{52} & \ding{56} & 32.0335 / 0.9203\\
        6 & \ding{52} & \ding{52} & \ding{52} & \textbf{32.0594} / \textbf{0.9205}\\
        \hline
        \end{tabular}
        \label{rdp}
\end{table}

\begin{figure}[t]
  \centering
  \includegraphics[width=1\linewidth]{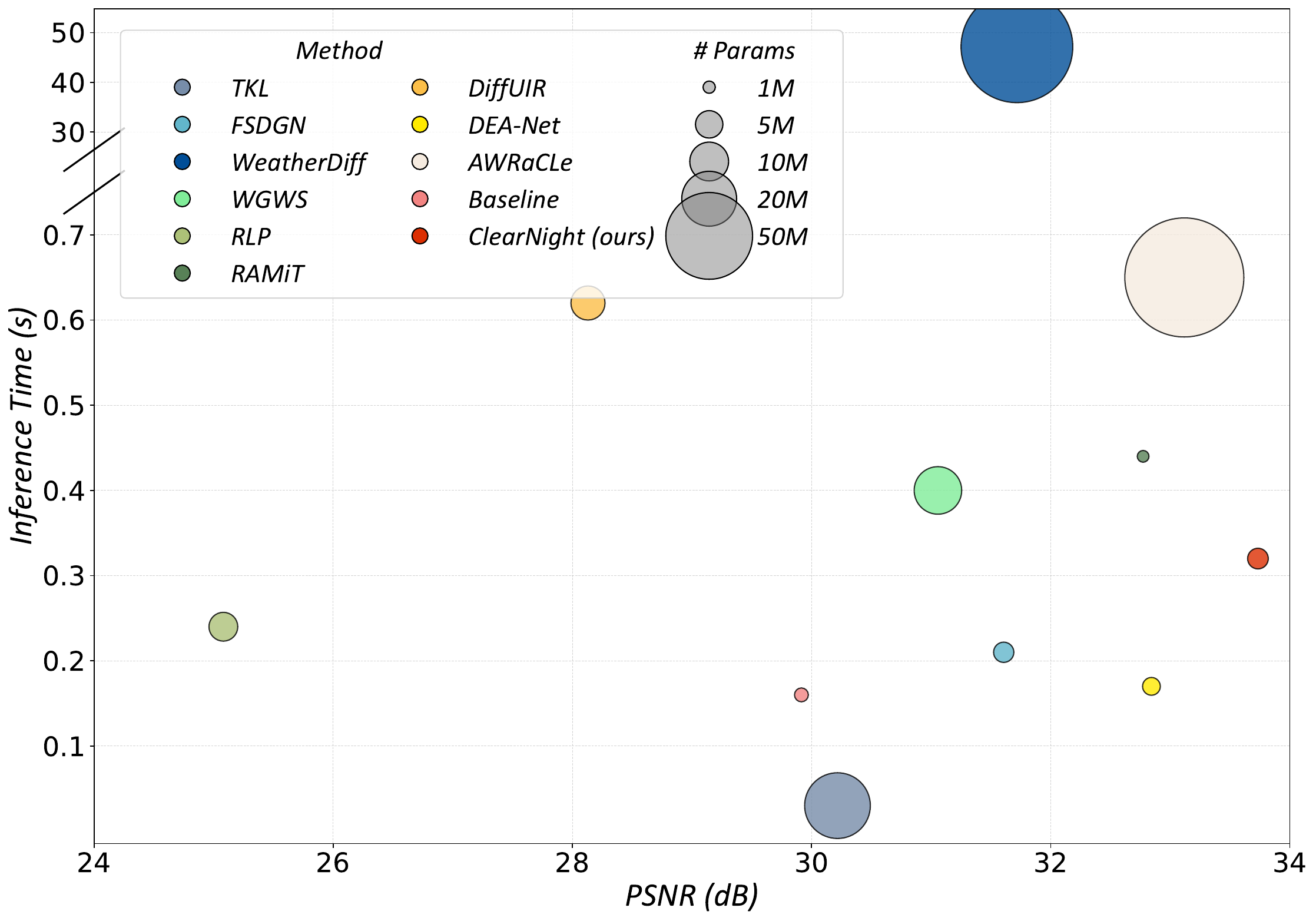}
   \caption{Comparison of ClearNight with other methods on AllWeatherNight synthetic testing subset. The marker size reflects the number of model parameters.}
   \label{complexity}
\end{figure}

\begin{figure*}[htbp]
  \centering
  \includegraphics[width=\linewidth]{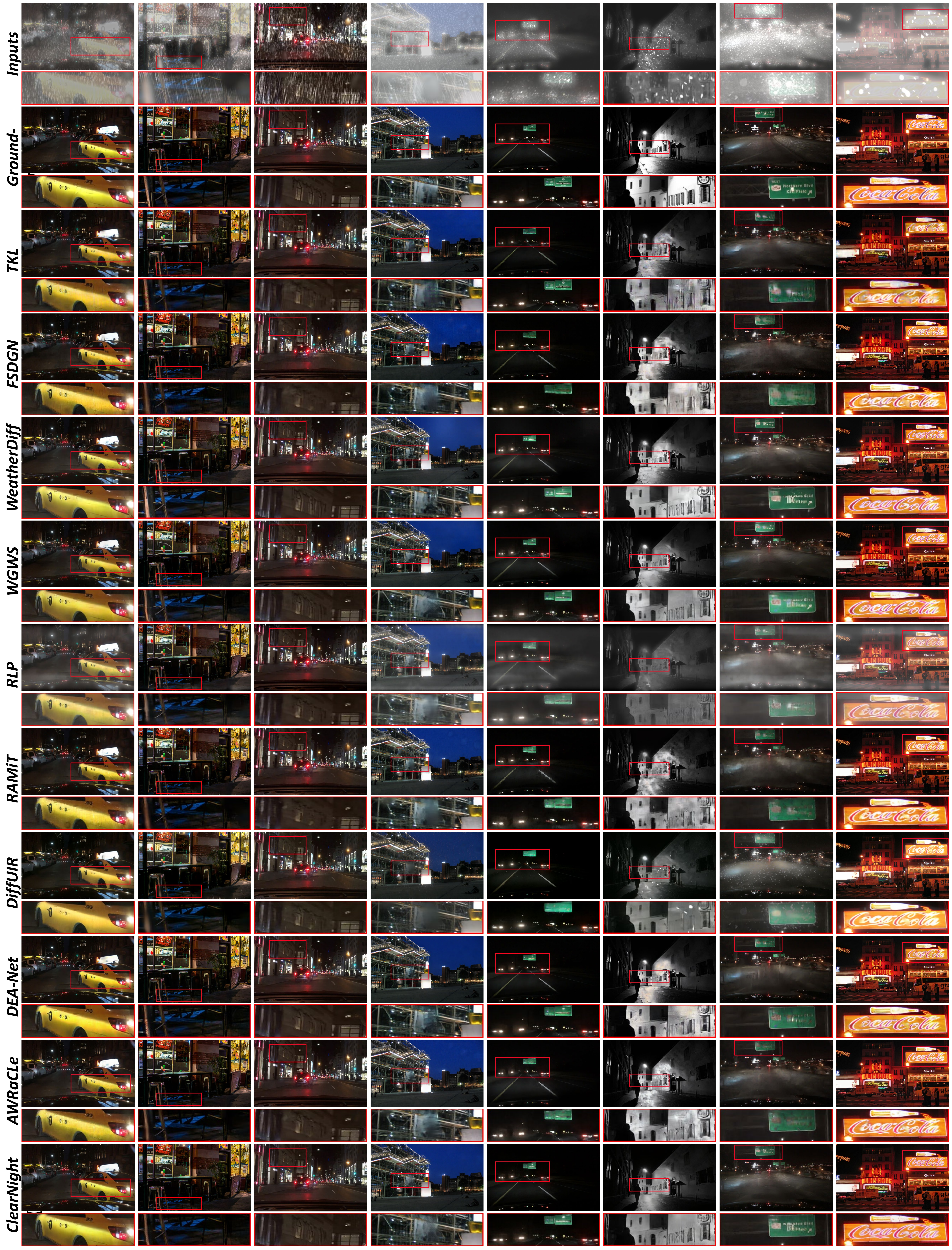}
   \caption{Qualitative results of rain and snow scenes in the AllWeatherNight.}
   \label{multi-result}
\end{figure*}

\begin{figure*}[htbp]
  \centering
  \includegraphics[width=0.725\linewidth]{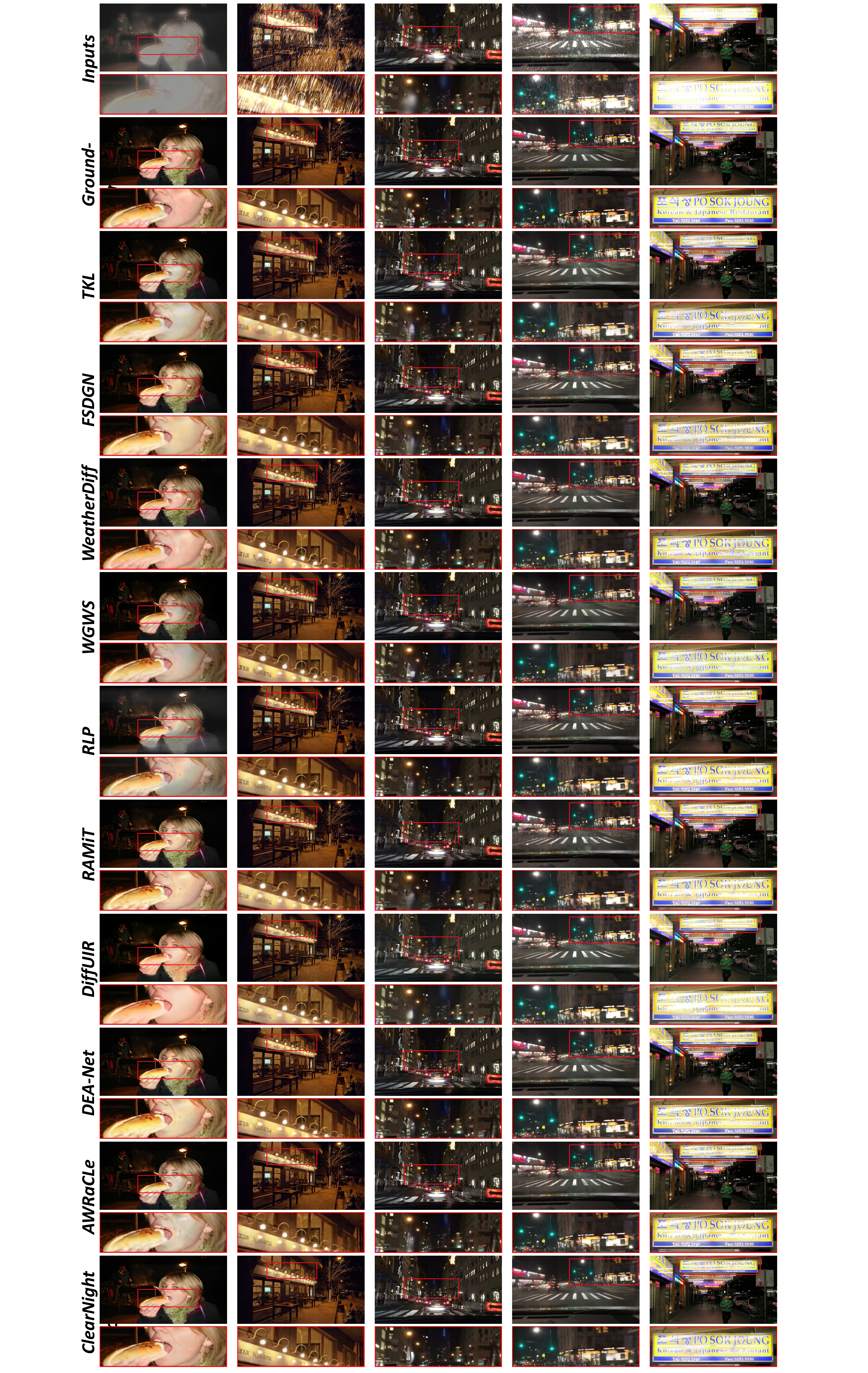}
   \caption{Qualitative results of single degradation in the AllWeatherNight, including haze, rain streak, raindrop, snow and flare.}
   \label{single_result}
\end{figure*}

\begin{figure*}[htbp]
  \centering
  \includegraphics[width=1\linewidth]{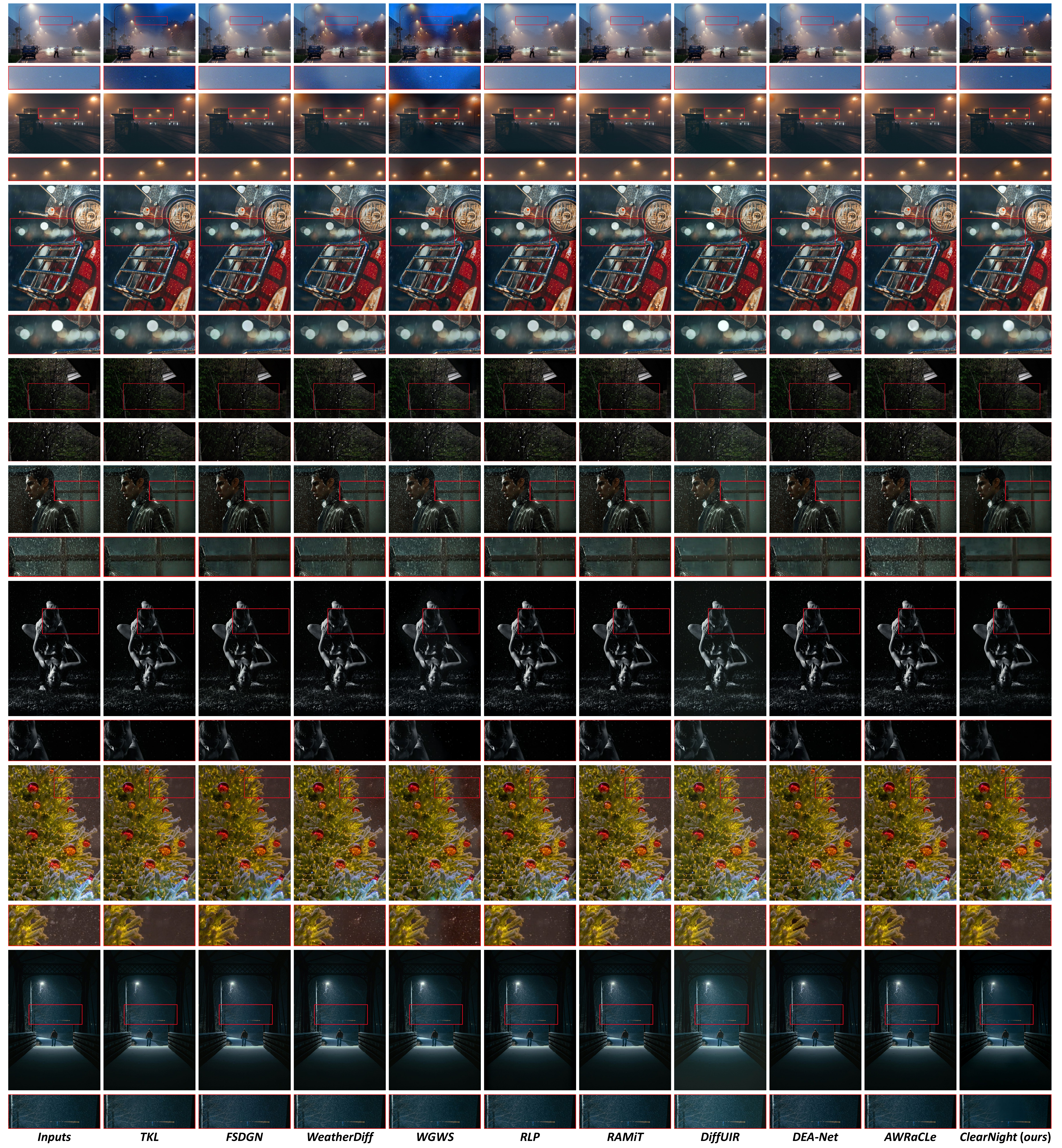}
   \caption{Qualitative results on real-world samples.}
   \label{real_result2}
\end{figure*}

\noindent\textbf{Analysis of the Loss Functions.} Our method employs the loss functions $\mathcal{L}_\mathrm{lb}$ and $\mathcal{L}_\mathrm{b}$ to supervise the dynamic allocator and weather instructor, respectively, enabling effective auxiliary selection tasks.
To evaluate their impact, we conduct experiments on AllWeatherNight, with results visualized in Fig.~\ref{two losses}. 
Evidently, $\mathcal{L}_\mathrm{lb}$ optimizes the allocation of candidate units, ensuring efficient utilization. 
Without $\mathcal{L}_\mathrm{lb}$, many units remain underutilized, degrading performance in complex multi-weather scenes.
Similarly, applying $\mathcal{L}_\mathrm{b}$ enhances the correlation between diverse weather conditions and candidate units selection, improving the model's ability to address multiple adverse weather effects. Without $\mathcal{L}_\mathrm{b}$, the selected units for different weather distortions tend to converge, impairing multi-weather nighttime image restoration. 
In contrast, the combined use of $\mathcal{L}_\mathrm{b}$ and $\mathcal{L}_\mathrm{lb}$ enables precise identification of different adverse weather degradations and balanced utilization of candidate units. 

As depicted in the bottom-right of Fig.~\ref{two losses}, the correlation visualization reveals that distinct units are selected for different adverse weather conditions, with similar sets of units consistently chosen for the same weather type, and all units are effectively utilized, maintaining balanced load distribution. 
Consequently, this dual-loss collaboration strategy significantly enhances the model's performance in restoring complex nighttime scenarios with diverse adverse weather conditions and flare effects, achieving high-quality results. 

The weighting of the loss function critically influences the performance of the network during training, particularly for multi-weather nighttime scenes with flare effects. 
To determine an optimal weighting strategy for $\mathcal{L}_\mathrm{b}$, we conduct an ablation study on the Rain Scene in AllWeatherNight dataset, with results shown in Fig.~\ref{bce}.
The absence of $\mathcal{L}_\mathrm{b}$ significantly degrades model's performance, as it fails to distinguish diverse adverse weather degradation features, impairing the guidance for unit selection.
When the weight was set to 0.01, the excessively strong constraint of $\mathcal{L}_\mathrm{b}$ limits its restoration performance on different adverse weather degradations with similar patterns. 
Conversely, with an excessively low weight of 0.0001, various adverse weather features cannot be effectively distinguished, resulting in poor performance across multiple weather conditions.
Weights of 0.001 achieve the best performance. Therefore, we set the $\lambda_\mathrm{b}$ to 0.001 for our model.

\subsection{Inference Efficiency}\label{sec: CC}
Fig.~\ref{complexity} compares the average parameters and inference time of our method with recent competitive approaches across all synthetic scenes. While diffusion-based method such as WeatherDiff~\cite{WeatherDiff2023} and DiffUIR~\cite{SCVPR2024} achieve impressive performance, they suffer from higher inference time, which limits their practicality in real-world scenes. On the other hand, lightweight models such as RAMiT~\cite{RAMITCVPR2024} and DEA-Net~\cite{TIP2024DEA} demonstrate fast inference, but often compromise restoration quality. 
Our ClearNight achieves a good trade-off between performance and efficiency. With only 2.84M parameters and an inference time of 0.32s, it is significantly faster than most diffusion-based models while offering much stronger restoration performance than lightweight models.
In summary, ClearNight maintains a relatively low computational cost, making it a practical and effective solution for multi-weather nighttime image restoration.

\subsection{Additional Qualitative Comparisons}\label{sec: results}
\noindent\textbf{Multi-Degradation.} We have provided more visual results of our model on Rain Scene and Snow Scene in AllWeatherNight, as shown in Fig.~\ref{multi-result}. 
The Rain and Snow Scene testing subsets contain degraded images with complex multi-weather degradations, effectively simulating real-world nighttime scenes. 
The visual results demonstrate that ClearNight excels in removing the influence of adverse weather while restoring rich background details. 
Especially when the background is similar to the adverse weather degradation, ClearNight accurately distinguishes these factors, producing high-quality images.

\noindent\textbf{Single-Degradation.} Although multiple adverse weather conditions frequently occur in an intertwined manner, the challenge of restoring scenes under simple degradations remains a crucial and non-trivial aspect of our work. Thus, we evaluate the generalization capability of ClearNight on single-degradation conditions. As shown in Fig.~\ref{single_result}, we can see that our method is equally effective in single-degradation scenarios. 

\noindent\textbf{Real-World.} Nighttime real-world degraded images often contain complex weather interferences and diverse flares. As shown in Fig.~\ref{real_result2}, we provide more comparison result of real-world scenes, \textit{e.g.}, haze ($1^\mathrm{st}$ and $2^\mathrm{nd}$ rows), rain ($3^\mathrm{rd}$ and $4^\mathrm{th}$ rows), raindrop ($5^\mathrm{th}$ and $6^\mathrm{th}$ rows), snow ($7^\mathrm{th}$ and $8^\mathrm{th}$ rows). The rain and snow scenes often exhibit subtle haze effects, as evidenced by real-world nighttime images, consistent with multi-degradation scenarios in our dataset.
The restored results show that ClearNight effectively removes the interference of multiple weather conditions while recovering natural nighttime lighting. 
Compared to existing adverse weather image restoration approaches, our method is superior in reconstructing scene content and natural lighting conditions.

\bibliography{aaai2026}

\end{document}